\documentclass[letterpaper]{article} 
\usepackage{aaai2026}  
\usepackage{times}  
\usepackage{helvet}  
\usepackage{courier}  
\usepackage[hyphens]{url}  
\usepackage{graphicx} 
\usepackage{natbib}  
\usepackage{caption} 
\usepackage{algorithm}
\usepackage{algorithmic}

\usepackage{newfloat}
\usepackage{listings}
\DeclareCaptionStyle{ruled}{labelfont=normalfont,labelsep=colon,strut=off} 
\floatstyle{ruled}
\newfloat{listing}{tb}{lst}{}
\floatname{listing}{Listing}
\usepackage{xspace}
\newcommand{\eg}{\textit{e.g.}\@\xspace}

\usepackage[T1]{fontenc}
\usepackage{amsmath}
\usepackage{amsfonts}
\usepackage{booktabs}
\usepackage{multirow}
\usepackage{appendix}

\usepackage[]{color-edits}
\addauthor[suppress]{hj}{blue}
\addauthor[suppress]{jhy}{red}
\usepackage{cleveref}
\Crefname{figure}{Fig.}{Figs.}

\title{Phased One-Step Adversarial Equilibrium for Video Diffusion Models}
\author {
    Jiaxiang Cheng\textsuperscript{\rm 1},
    Bing Ma\textsuperscript{\rm 1},
    Xuhua Ren\textsuperscript{\rm 1$\ast$},
    Hongyi Henry Jin\textsuperscript{\rm 1,2},\\
    Kai Yu\textsuperscript{\rm 1},
    Peng Zhang\textsuperscript{\rm 1},
    Wenyue Li\textsuperscript{\rm 1},
    Yuan Zhou\textsuperscript{\rm 1},
    Tianxiang Zheng\textsuperscript{\rm 1},
    Qinglin Lu\textsuperscript{\rm 1}\thanks{Corresponding authors.}
}
\affiliations {
    \textsuperscript{\rm 1}Tencent Hunyuan\\
    \textsuperscript{\rm 2}Computer Science Department, UCLA \\
    jiaxiangcc@gmail.com
}

\begin{document}

\maketitle

\begin{abstract}
Video diffusion generation suffers from critical sampling efficiency bottlenecks, particularly for large-scale models and long contexts.
Existing video acceleration methods, adapted from image-based techniques, lack a single-step distillation ability for large-scale video models and task generalization for conditional downstream tasks.
To bridge this gap, we propose the Video Phased Adversarial Equilibrium (\textit{V-PAE}), a distillation framework that enables high-quality, single-step video generation from large-scale video models. Our approach employs a two-phase process.
(i) Stability priming is a warm-up process to align the distributions of real and generated videos. It improves the stability of single-step adversarial distillation in the following process.
(ii) Unified adversarial equilibrium is a flexible self-adversarial process that reuses generator parameters for the discriminator backbone. It achieves a co-evolutionary adversarial equilibrium in the Gaussian noise space.
For the conditional tasks, we primarily preserve video-image subject consistency, which is caused by semantic degradation and conditional frame collapse during the distillation training in image-to-video (I2V) generation.
Comprehensive experiments on VBench-I2V demonstrate that V-PAE outperforms existing acceleration methods by an average of 5.8\% in the overall quality score, including semantic alignment, temporal coherence, and frame quality.
In addition, our approach reduces the diffusion latency of the large-scale video model (\eg, Wan2.1-I2V-14B) by $100\times$, while preserving competitive performance.
\end{abstract}

\begin{links}
    \link{Project Page}{https://v-pae.github.io/}
\end{links}

\begin{figure}[tb]
  \centering
  \includegraphics[width=1.0\columnwidth]{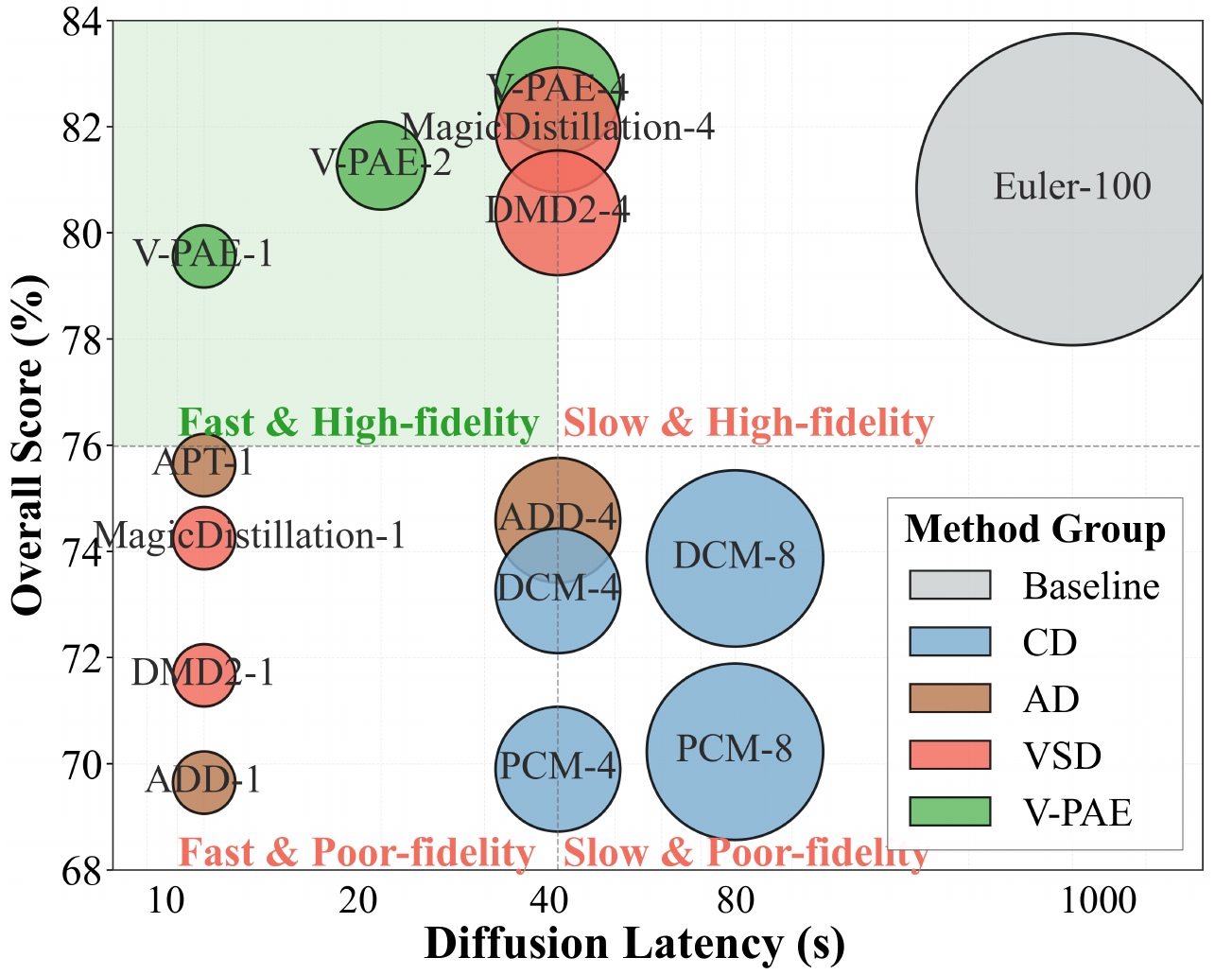} 
  \caption{
\textbf{Comparison between V-PAE and existing acceleration methods on VBench-I2V}. It includes three distillation paradigms: (i) Consistency Distillation (CD), (ii) Variational Score Distillation (VSD) and (iii) Adversarial Distillation (AD). For fairness, all models are distilled from Wan2.1-I2V-14B~\cite{wan2025wan} using the same dataset and training cost. Diffusion latency is measured for 5-second $720 \times 1280$ videos on $8 \times$ H20 GPUs.
  }
  \label{fig:teaser}
\end{figure}

\section{Introduction}
\label{sec:intro}

Diffusion models~\cite{songscore,ho2020denoising,songdenoising} have witnessed a paradigm shift, extending their remarkable success from image synthesis~\cite{rombach2022high,podellsdxl,cheng2025resadapter,chen2024pixart,esser2024scaling} to video generation~\cite{wang2023modelscope,guoanimatediff,blattmann2023stable,yang2024cogvideox,kong2024hunyuanvideo,wan2025wan}. 
Contemporary video diffusion models attain impressive fidelity through two factors: architectural scaling and longer temporal context. Designs range from the small-scale U-Net~\cite{ronneberger2015u} employing spatiotemporal decoupled attention to the large-scale Diffusion Transformer (DiT)~\cite{peebles2023scalable} employing full attention, while temporal windows extend to hundreds of frames.
However, this fidelity comes at a prohibitive computational cost.
For instance, synthesizing a 5-second video with 50 diffusion steps using a large-scale video model such as Wan2.1-I2V-14B\footnote{https://huggingface.co/Wan-AI/Wan2.1-I2V-14B-720P}~\cite{wan2025wan} requires approximately 15 minutes on advanced GPUs (\eg, 8$\times$H20).
This substantial latency severely limits deployment in real-time applications.

Existing video diffusion distillation methods~\cite{wang2023videolcm,lit2v,zhang2025accvideo,lin2025diffusion,shao2025magicdistillation} primarily adapt image distillation techniques~\cite{salimansprogressive,song2023consistency,luo2023latent,yin2024one,sauer2024adversarial,sauer2024fast,yin2024improved}, lacking critical spatiotemporal complexities for video data.
We identify two fundamental limitations:
(i) efficiency bottleneck, failing to distill large-scale video diffusion models ($>$10B) into single-step generators;
and (ii) task generalization, lacking the ability to perform conditional tasks (\eg, image-to-video generation (I2V)). 
It suffers from poor video-image subject consistency caused by semantic degradation and conditional frame collapse.

To address the efficiency bottleneck, our objective is to distill a model capable of generating single-step videos that closely match real videos.
Adversarial distillation for single-step sampling from Gaussian noise is the dominant strategy. But it faces severe challenges: the significant quality gap between real and generated videos makes discrimination trivial, yielding weak gradients and unstable training.
As a result, existing adversarial methods, such as Distribution Matching Distillation (DMD2)~\cite{yin2024improved} and adversarial diffusion distillation (ADD)~\cite{sauer2024fast}, begin denoising from medium-to-high signal-to-noise ratio (SNR) engines to ensure sufficient similarity between real and generated samples, which provides informative discriminator supervision.
However, the mismatch between these training distributions and low-SNR sampling fundamentally limits the performance of the distilled model.
To this end, we propose Video Phased Adversarial Equilibrium (\textbf{V-PAE}), which comprises two sequential processes: stability priming and unified adversarial equilibrium.
(i) Stability priming is a warm-up process to align the distributions of real and generated videos, which improves the stability of single-step adversarial distillation in the subsequent process.
Guided by Variational Score Distillation (VSD)~\cite{pooledreamfusion}, it relies on score-gradient differences to reduce the distributional gap.
(ii) Unified adversarial equilibrium is a flexible self-adversarial process that reuses generator parameters for the discriminator backbone. It achieves a co-evolutionary adversarial equilibrium in the Gaussian noise space,  improving the single-step adversarial stability.

To improve task generalization, we introduce the semantic discriminator head and the conditional Score Distillation Sampling (SDS)~\cite{pooledreamfusion} loss.
The semantic discriminator head primarily enhances the semantic perception of images and videos. It employs the multi-modal (\eg, image, video, and text) attention module. 
By enabling cross-modal interactions with learnable query tokens, it strengthens the semantic alignment ability of the distilled video model.
The conditional SDS is inspired by ADD~\cite{sauer2024adversarial}, leveraging the continuous distribution of the pretrained video model to prevent distribution bias.
It requires only a multi-step video model as the regularization distribution, correcting conditional frame collapse based on smaller discrepancies between the conditional and generated frames.

We comprehensively evaluate V-PAE on the VBench-I2V benchmark~\cite{huang2024vbench} for semantic alignment, temporal coherence, and frame quality.
Experiments demonstrate that our approach outperforms other acceleration methods by an average of 5.8\% in the overall quality score of single-step videos.
Crucially, our approach achieves single-step quality comparable to that of its 50-step video model and even exceeds it with a few steps in a zero-shot manner.
As illustrated in Fig.~\ref{fig:teaser}, applied to the large-scale video diffusion model (\eg, Wan2.1-I2V-14B~\cite{wan2025wan}), our approach reduces the iterative time from nearly 15 minutes to 10 seconds, achieving a $100\times$ speedup.
This overcomes the computational barrier, enabling real-time, high-fidelity video synthesis for interactive applications.


\section{Related Work}
\label{sec:related}

\subsubsection{Video Diffusion Models}
\label{sec:video-diffusion-models}

The video generation field has seen rapid progress through diffusion models~\cite{songscore,ho2020denoising,songdenoising}, enabling applications such as conditional generation~\cite{hacohen2024ltx,kong2024hunyuanvideo,wan2025wan,yang2024cogvideox} and video editing~\cite{chai2023stablevideo,feng2024ccedit,singer2024video}.
Early works~\cite{wang2023modelscope,guoanimatediff,blattmann2023stable} insert a lightweight temporal attention into U-Net~\cite{ronneberger2015u}, using spatiotemporal decoupling to model video distribution. 
Later works inherit weights from large-scale image diffusion models~\cite{peebles2023scalable,esser2024scaling,chen2024pixart} for video synthesis. 
Recent works~\cite{yang2024cogvideox,hacohen2024ltx,kong2024hunyuanvideo,wan2025wan} abandon reliance on image diffusion models. Instead, they build comprehensive video synthesis systems through data expansion, model scaling, and training optimization procedures. 
However, they remain limited in computational efficiency and inference latency, primarily due to the large-scale model parameters and long-video temporal complexity.


\subsubsection{Distillation Sampling}
\label{sec:distillation-sampling}

Reducing diffusion sampling steps remains a central challenge.
In image distillation, Progressive Distillation (PD) \cite{salimansprogressive,lin2024sdxl} progressively compresses teacher-student pipelines but faces inherent performance limits due to error accumulation.
Consistency Distillation (CD) \cite{song2023consistency,luo2023latent} enables single-step generation along the Probability Flow (PF) Ordinary Differential Equation (ODE); however, training stability remains an issue.
Variational Score Distillation (VSD) \cite{yin2024one,yin2024improved} minimizes gradient differences between real and fake score estimates; however, fidelity constraints persist in single-step sampling.
Adversarial Distillation (AD) \cite{sauer2024adversarial,lin2024sdxl,sauer2024fast} uses a discriminator to reduce the distribution gap between real and generated data; however, effective distillation is limited to four uniformly spaced timesteps.
In video distillation, most works simply adapt image techniques.
Dual-Expert Consistency Model (DCM) \cite{wang2023videolcm,lit2v,lv2025dcm} applies consistency distillation without video-specific designs.
MagicDistillation (MD) \cite{shao2025magicdistillation} achieves few-step synthesis via weak-to-strong score matching, but the quality degrades at single-step sampling.
Adversarial Post-Training (APT)~\cite{lin2025diffusion} deploys one-step adversarial distillation for videos, but its applicability is restricted to small-scale models and short clips.
\begin{figure*}[tb]
  \centering
  \includegraphics[width=0.95\linewidth]{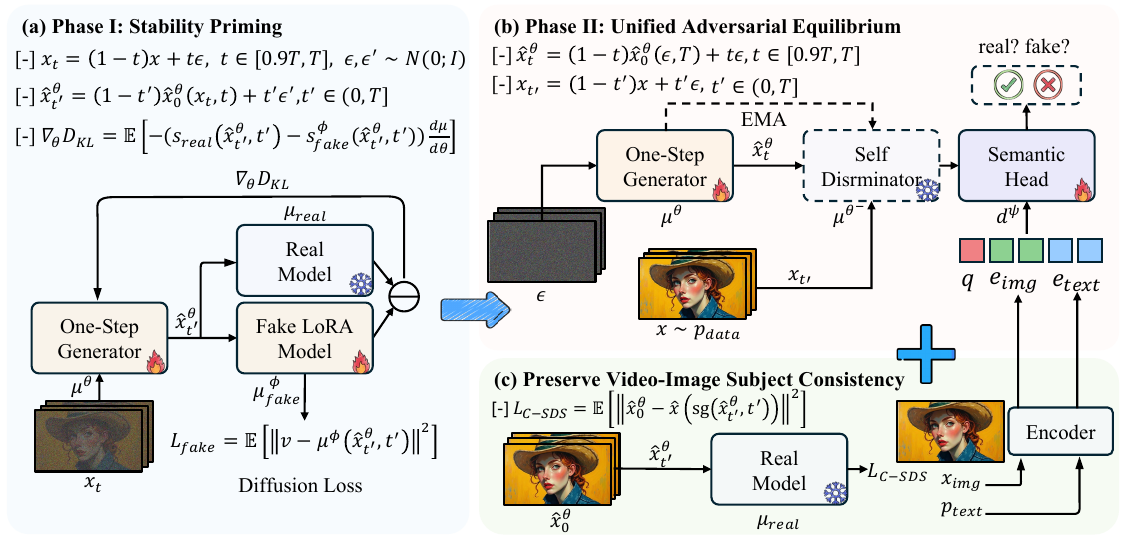}  
  \caption{
\textbf{Overview.}
\textit{V-PAE} first aligns the distributions of generated and real videos in the (a) \textit{stability priming} process.
Building on this process, it reuses the generator parameters for the discriminator backbone, which achieves a co-evolutionary adversarial training in the (b) \textit{unified adversarial equilibrium} process.
For the conditional generation, we also provide the conditional SDS loss and semantic discriminator to (c) \textit{preserve video-image subject consistency}.
  }
  \label{fig:overview}
\end{figure*}

\section{Methodology}
\label{sec:method}

Our objective is to distill video diffusion models that can generate single-step video $\hat{x}^\theta_0$ close to real video $x\sim p_\text{data}$.
To this end, we propose V-PAE, a two-phase, one-step equilibrium distillation framework.
First, we introduce a stability priming process in Sec.~\ref{sec:stability-priming}, which  is a warm-up process to align the distributions between $\hat{x}^\theta_0$ and $x$. 
It improves the stability of single-step adversarial distillation in the following process.
Second, we introduce the unified adversarial equilibrium process in Sec.~\ref{sec:unified-adversarial-equilibrium}, which is a flexible self-adversarial process that reuses generator parameters for the discriminator backbone. 
It achieves a co-evolutionary equilibrium.
For the conditional tasks, such as image-to-video (I2V) generation, we primarily preserve the video-image subject consistency in Sec.~\ref{sec:conditional-adversarial-consistency}, which avoids semantic degradation and conditional frame collapse.
Finally, we provide the total training optimization objective in Sec~\ref{sec:traing-objective}.

\subsection{Phase I: Stability Priming}
\label{sec:stability-priming}
Adversarial distillation faces severe training instability for the single-step video $\hat{x}^\theta_0$ generated from Gaussian noise $\epsilon$. It is caused by the distribution mismatch between the $\hat{x}^\theta_0$ and the real video $x \sim p_\text{data}$.
To this end, we design a stability priming process to narrow their distribution distance.
We define three models:  a priming generator $\mu^\theta$ to align the single-step distribution with the real distribution, a real model $\mu_\text{real}$, and a fake model $\mu^\phi_\text{fake}$ for score-based estimates of distribution matching distillation.
As illustrated in Fig.~\ref{fig:overview} (a), the $\mu^\theta$ directly generates $\hat{x}^\theta_0=f^\theta(x_t,t)$ in the low signal-to-noise ratio (SNR) regime, where $t\in[0.9T,T]$.
The $\hat{x}^\theta_0$ is perturbed by varying Gaussian noise $\epsilon$ to obtain $\hat{x}^\theta_{t'}=(1-t')\hat{x}^\theta_0+t'\epsilon$, which inputs $\mu_\text{real}$ and $\mu^\phi_\text{fake}$ to compute the distribution matching loss with the score gradient difference, as shown in Eq.~\ref{eq:dmd_kl}:
\begin{equation}
    \nabla_\theta D_\text{KL} = \underset{\underset{}{\epsilon \sim \mathcal{N}(0;\textbf{I})}}{\mathbb{E}} \left[ - (s_\text{real}(\hat{x}^\theta_{t'}) - s^\phi_\text{fake}(\hat{x}^\theta_{t'})) \frac{d \mu}{d \theta} \right],
    \label{eq:dmd_kl}
\end{equation}
where $t'\in (0,T]$, $s_\text{real} = - \frac{\hat{x}^\theta_{t'} + (1-t') \mu_\text{real}}{t'}$, and $s_\text{fake} = - \frac{\hat{x}^\theta_{t
'}+ (1-t') \mu_\text{fake}^\phi}{t'}$ are under standard score-based definitions~\cite{songscore}.
\subsubsection{Efficient Distribution Track}
The real score is fixed during training as the real distribution.
The fake score is dynamic during training as the single-step distribution, which is updated with the standard diffusion loss~\cite{ho2020denoising} as shown in Eq.~\ref{eq:fake_loss}.
\begin{equation}
    \mathcal{L}_\text{fake} = \underset{\underset{}{\epsilon \sim \mathcal{N}(0;\textbf{I})}}{\mathbb{E}}  [ ||v - \mu_\text{fake}^\phi (\hat{x}^\theta_{t'}, t')  ||^ 2 ],
    \label{eq:fake_loss}
\end{equation}
To improve the stability and efficiency of tracking the distribution of the $\hat{x}^\theta_0$, we introduce the lightweight Low-Rank Adaptation (LoRA)~\cite{hulora} to $\mu^\phi_\text{fake}$.
Compared to full-parameter training, this strategy enables a more stable and faster distribution tracking ability for the large-scale video model, even with only a small amount $\hat{x}^\theta_0$.
For stable launching, we additionally enable zero-parameter initialization in the adaptation weights.

\begin{figure}[tb]
  \centering
  \includegraphics[width=0.9\columnwidth]{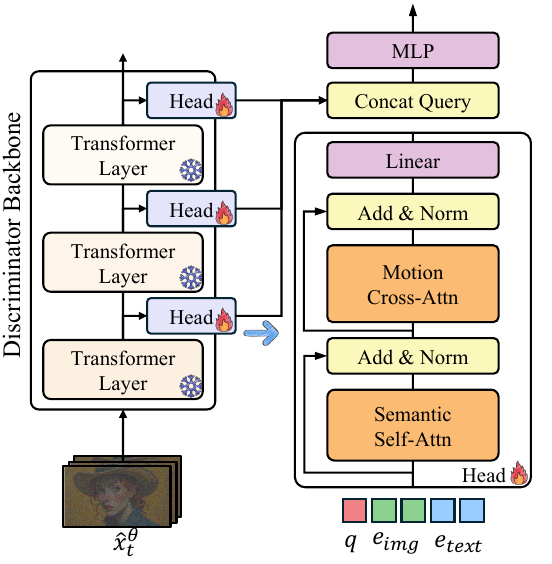}  
  \caption{
    The semantic discriminator head architecture.
  }
  \label{fig:head}
\end{figure}

\subsection{Phase II: Unified Adversarial Equilibrium}
\label{sec:unified-adversarial-equilibrium}

Building on the stability priming process, we propose the unified adversarial equilibrium process to guide the $\mu^\theta$ in generating high-quality single-step videos $\hat{x}^\theta_0$ from Gaussian noise $\epsilon$.
In traditional adversarial distillation~\cite{sauer2024adversarial,sauer2024fast}, the discriminator backbone is either frozen or fully-parameter-trained.
The frozen backbone easily leads to significant asymmetry in the trained parameters  between the generator and discriminator. 
It tends to degrade the overall quality of the single-step video.
The trained backbone faces the memory challenge for the large-scale video diffusion model (\eg, Wan2.1-I2V-14B), as shown in Tab.~\ref{tab:ablation}.

\subsubsection{Self Discriminator}
Instead, we introduce a flexible adversarial paradigm.
It enables the $\mu^\theta$ as a self discriminator backbone, paired with lightweight heads $d^\psi$ for computing logits.
The key property is efficient co-evolution.
With the limited memory, the self discriminator backbone maintains a comparable learning capacity. It enables a stable and high-quality Nash equilibrium while generating the single-step video from Gaussian noise.
As illustrated in Fig.~\ref{fig:overview} (b), the $\mu^\theta$ first samples $\hat{x}^\theta_0=f^\theta(\epsilon,T)$ from the endpoint.
Then $\hat{x}^\theta_0$ is perturbed with varying noise levels in low-SNR regimes.
The noisy samples $\hat{x}^\theta_{t}$  are fed into the self discriminator backbone $\mu^\theta$ to extract multi-layer features, which are used to compute logits by $d^\psi$.
Following the Hinge loss~\cite{miyato2018spectral}, our unified adversarial loss is as shown in Eq.~\ref{eq:adv_loss}:
\begin{equation}
\begin{aligned}
\mathcal{L}_{\text{UAE}} = 
& \underbrace{\mathbb{E}_{\hat{x}^\theta_0=f^\theta(\epsilon,T)} \left[ \log D^{\psi}_{\theta} \left(\hat{x}^\theta_t,t; \mu^\theta \right) \right]}_{\text{Generator objective}} \\
& +\underbrace{\mathbb{E}_{x \sim p_\text{data}} \left[ \log \left(1 - D_{\theta^-}^{\psi} \left(x_{t},t; \mu_{\text{}}^{\theta^-} \right)\right) \right]}_{\text{Discriminator objective}},  
\end{aligned}
\label{eq:adv_loss}
\end{equation}
where $D^{\psi}_{\theta^-}=\left[ \mu_\text{}^{\theta^-};d^\psi \right]$ is trained using the Exponential Moving Average (EMA) weights $\theta^-$ for unified adversarial stability, and the $d^\psi$ represents the discriminator heads.
\subsubsection{Gradient Regularization}
To mitigate the gradient explosion of the discriminator heads, we further introduce a spatiotemporal differential R1 regularization.
It treats temporal dimensions as intrinsic properties of the video, which enforce structured perturbations that jointly constrain spatial and temporal quality.
Specifically, we apply composite noise perturbations to the single-step video as $\tilde{x}^\theta_t=\hat{x}^\theta_t+\sigma_s \epsilon_s + \sigma_t \epsilon_t$. 
The gradient regularization for the video model is as shown in Eq~\ref{eq:str1_loss}:
\begin{equation}
    \mathcal{L}_{\text{STR1}} = \mathbb{E}_{\hat{x}^\theta_0=f(\epsilon,T)} \left[ \frac{ \|  D_{\theta^-}^{\psi}(\tilde{x}^\theta_{t}, t) - D_{\theta^-}^{\psi}(\hat{x}^\theta_t, t) \|^2 }{ \|{\sigma_s\epsilon}_s + {\sigma_t\epsilon}_t\|^2 } \right],
    \label{eq:str1_loss}
\end{equation}
where $\sigma_s=0.01$ is for perturbing the pixel distribution per frame, and $\sigma_t=0.1$ is for disturbing the temporal distribution across the frames.

\subsection{Preserve Video-Image Subject Consistency}
\label{sec:conditional-adversarial-consistency}
Compared to text-to-video (T2V) generation, image-to-video (I2V) generation is the primary video application.
However, its primary challenge is that the score of video-image subject consistency significantly degrades during the adversarial distillation process.
This phenomenon is caused by semantic degradation and conditional frame collapse, which is as shown in Fig.~\ref{fig:ablation}.
To this end, we introduce the semantic discriminator head and conditional Score Distillation Sampling (SDS) loss.

\subsubsection{Semantic Discriminator Head}
To improve semantic alignment, our discriminator heads integrate multi-modal information (\eg, video, image, and text).
Inspired by Adversarial Post-Training (APT)~\cite{lin2025diffusion}, we introduce a learnable logit query $q \in \mathbb{R}^{n \times d}$. 
As illustrated in Fig.~\ref{fig:head}, the $q$ concatenates with the conditional image embeds $e_\text{img} \in \mathbb{R}^{s \times d}$ and the text embeds $e_\text{text} \in \mathbb{R}^{s \times d}$, which fuse the semantic information. To distinguish between different modalities, we incorporate modality position indices.
After the semantic self-attention module, the  multi-layer features from the self discriminator backbone $\mu^\theta$ are cross-attended with the $q$, which enhances sensitivity to temporal quality.
Finally, all processed queries are concatenated along the channel dimension and projected to a scalar logit by a Feed-Forward Network (FFN) layer.

\subsubsection{Conditional SDS Loss}
To prevent the conditional frame collapse, the conditional SDS loss leverages the distributional stability of the $\mu_\text{real}$ to reduce the discrepancy between continuous frames.
As illustrated in Fig.~\ref{fig:overview} (c), the $\mu_\text{real}$ can maintain the consistency between the conditional and generated frames. 
Inspired by SDS~\cite{pooledreamfusion}, this loss can be defined as shown in Eq.~\ref{eq:frame_loss}:
\begin{equation}
\mathcal{L}_\text{C-SDS}=\mathbb{E}_{\hat{x}^\theta_0} \left[ \lVert \hat{x}_0^\theta - f_\text{real}(\text{sg}(\hat{x}_{t'}^\theta),t') \rVert^2 \right],
    \label{eq:frame_loss}
\end{equation}
Unlike the poor effect in image adversarial distillation~\cite{sauer2024adversarial,sauer2024fast}, this mechanism significantly avoids disturbances in the conditional frame.

\begin{figure*}[htb]
  \centering
  \includegraphics[width=0.95\linewidth]{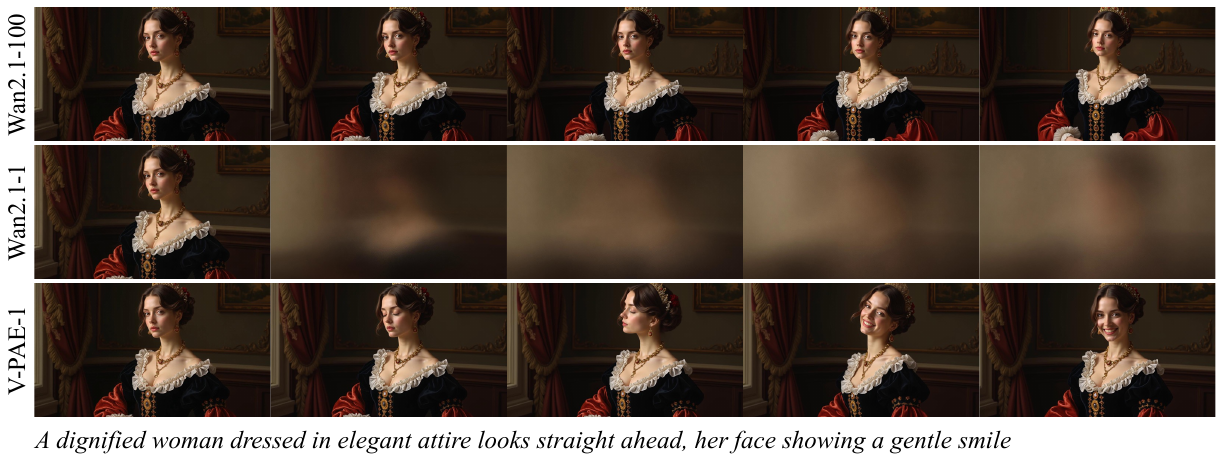}
  \caption{
\textbf{Qualitative comparison with Wan2.1-I2V-14B.} 
We compare our method against the baseline using both 100-NFE and 1-NFE sampling. 
For 100-NFE, videos are generated with 50 denoising steps and a guidance scale of 5.0.
  }
  \label{fig:qualitative02}
\end{figure*}
\begin{table*}[htbp]
\centering
\large
\resizebox{0.95\linewidth}{!}{
\begin{tabular}{@{}l|c|c|c|c|c|c|c|c|c|c@{}}
\toprule
\multirow{3}{*}{\textbf{Methods}} &
  \multirow{3}{*}{\textbf{Type}} &
  \multirow{3}{*}{\begin{tabular}[c]{@{}c@{}}\textbf{I2V} \\ \textbf{Score} \end{tabular}} \multirow{3}{*}{($\uparrow$)} &
  \multicolumn{2}{c|}{\textbf{Semantic Alignment (SA)}} &
  \multicolumn{2}{c|}{\textbf{Temporal Coherence (TC)}} &
  \multicolumn{2}{c|}{\textbf{Frame Quality (FQ)}} &
  \multirow{3}{*}{\begin{tabular}[c]{@{}c@{}}\textbf{Quality} \\ \textbf{Score}\end{tabular}} \multirow{3}{*}{($\uparrow$)}&
  \multirow{3}{*}{\begin{tabular}[c]{@{}c@{}}\textbf{Latency}\\ \textbf{Time (s)}\end{tabular}} \\
 &  &  &
  \begin{tabular}[c]{@{}c@{}}\textbf{Subject} \\ \textbf{Consistency}\end{tabular} ($\uparrow$)&
  \begin{tabular}[c]{@{}c@{}}\textbf{Background} \\ \textbf{Consistency}\end{tabular} ($\uparrow$)&
  \begin{tabular}[c]{@{}c@{}}\textbf{Motion} \\ \textbf{Smoothness}\end{tabular} ($\uparrow$)&
  \begin{tabular}[c]{@{}c@{}}\textbf{Dynamic} \\ \textbf{Degree}\end{tabular} ($\uparrow$)&
  \begin{tabular}[c]{@{}c@{}}\textbf{Aesthetic} \\ \textbf{Quality}\end{tabular} ($\uparrow$)&
  \begin{tabular}[c]{@{}c@{}}\textbf{Image} \\ \textbf{Quality}\end{tabular} ($\uparrow$)&
   &
   \\ 
\midrule
\multicolumn{11}{c}{\textbf{\textit{100-NFE}}} \\
\midrule
Baseline               & Euler & 92.90 & 94.86 & 97.07 & 97.90 & 51.38 & 64.75 & 70.44 & 80.82 & 890.15  \\
\midrule
\multicolumn{11}{c}{\textbf{\textit{4-NFE}}} \\
\midrule
Baseline               & Euler & 75.97 & 79.59 & 81.24 & 80.92 & 42.7 & 54.34 & 57.55 & 64.47 & \multirow{8}{*}{37.48} \\
DMD2              & VSD   & 91.9 & 93.19 & 95.37 & 96.05 & \underline{50.61} & 64.9 & \underline{70.65} & 80.38 &  \\
MD & VSD   & \underline{94.24} & \textbf{96.83} & \underline{97.90}  & \textbf{99.76}  & 50.60 & \underline{65.4} & 69.82 & \underline{81.94} &  \\
LCM               & CD    & 79.79 & 81.21 & 83.87 & 81.39 & 43.86 & 54.18 & 58.26 & 68.94 &  \\
PCM               & CD    & 80.54 & 80.82 & 84.12 & 83.01 & 44.89 & 55.67 & 60.27 & 69.9 &  \\
DCM               & CD    & 83.01 & 85.49 & 87.27 & 87.74 & 46.25 & 59.33 & 63.75 & 73.26 &  \\
ADD               & AD    & 84.48 & 87.53 & 90.07 & 90.77 & 46.71 & 59.06 & 63.5 & 74.59 &  \\
\textbf{V-PAE}               & AD    & \textbf{94.93} & \underline{94.21} & \textbf{98.43} & \underline{98.64} & \textbf{52.14} & \textbf{65.73} & \textbf{70.76} & \textbf{82.24} &  \\
\midrule
\multicolumn{11}{c}{\textbf{\textit{1-NFE}}} \\
\midrule
Baseline               & Euler & 69.59 & 70.73 & 74.95 & 72.94 & 38.51 & 49.72 & 54.06 & 61.52 & \multirow{7}{*}{9.37} \\
DMD2              & VSD   & 83.15 & 81.91 & 85.47 & 85.87 & 45.78 & 57.04 & 62.47 & 71.67 &  \\
MD & VSD   & 84.02 & \underline{87.87} & 87.9 & 90.04 & 46.32 & 58.85 & \underline{64.76} & 74.25 &  \\
LCM               & CD    & 76.92 & 79.1 & 81.0 & 82.55 & 43.06 & 53.94 & 59.15 & 67.96 &  \\
ADD               & AD    & 79.38 & 82.4 & 83.83 & 83.01 & 44.15 & 54.63 & 60.12 & 69.65 &  \\
APT               & AD    & \underline{84.87} & 87.11 & \underline{88.77} & \underline{90.94} & \underline{48.29} & \underline{60.33} & 64.69 & \underline{75.21} &  \\
\textbf{V-PAE}              & AD    & \textbf{91.54} & \textbf{94.14} & \textbf{95.8} & \textbf{94.99} & \textbf{49.54} & \textbf{62.28} & \textbf{68.66} & \textbf{79.56} &  \\ \bottomrule
\end{tabular}
}
\caption{
\textbf{Quantitative results on VBench-I2V.} The baseline model is Wan2.1-I2V-14B~\cite{wan2025wan}.
For each metric, the best result is highlighted in \textbf{bold}, and the second best is \underline{underlined}. 
Various distillation methods are compared under different numbers of sampling steps. 
Latency time is measured on $8 \times$ H20 for 5-second videos with a resolution of $720 \times 1080$.
}
\label{tab:main_quantitative}
\end{table*}

\subsection{Total Training Objective}
\label{sec:traing-objective}
Based on the processes mentioned above, our training objective for the single-step generator can be defined as:
\begin{equation}
\mathcal{L}_\text{G} = \mathcal{L}_\text{UAE-G} + \lambda \mathcal{L}_\text{C-SDS},
    \label{eq:ob-g}
\end{equation}
which consists of the adversarial generative loss and the conditional SDS loss. Where the $\lambda=10$ represents the injection strength.
And our training objective for the discriminator heads can be defined as:
\begin{equation}
\mathcal{L}_\text{D} = \mathcal{L}_\text{UAE-D} + \mathcal{L}_\text{STR1}(\sigma_s,\sigma_t),
    \label{eq:ob-g}
\end{equation}
which consists of the adversarial discriminative loss and the gradient regularization loss. Where the $\sigma_s=0.01$ and $\sigma_t=0.1$ represent the disturbance strengths.

\begin{figure*}[htb]
  \centering
  \includegraphics[width=0.95\linewidth]{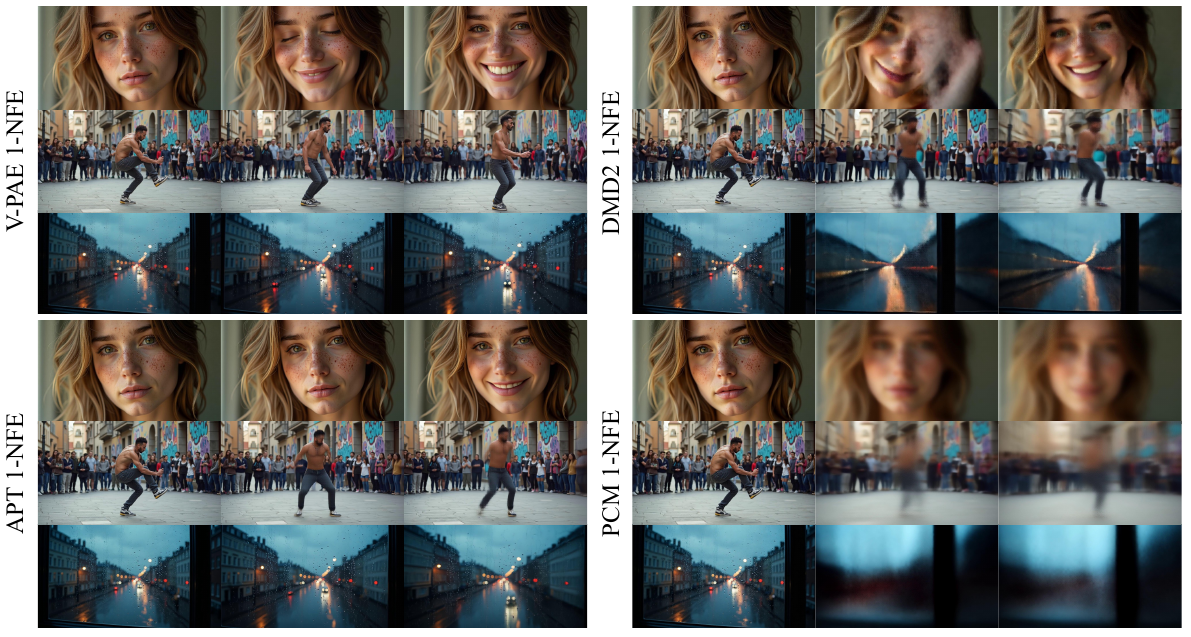}  
  \caption{
    \textbf{Qualitative results of 1-NFE between V-PAE and existing acceleration distillation methods.} 
    We evaluate against representative methods from three paradigms, including DMD2~\cite{yin2024improved} from VSD, PCM~\cite{wang2024phased} from CD, and APT~\cite{lin2025diffusion} from AD.
  }
  \label{fig:qualitative01}
\end{figure*}

\section{Experiment}
\label{sec:experiment}

We first present the detailed implementation for datasets, training, evaluation metrics, and baseline methods in Sec.~\ref{sec:setup} .
And we demonstrate the effectiveness of our approach through quantitative results and qualitative visualization comparisons in Sec.~\ref{sec:main-results}.
Then we analyze the necessity of our modules in Sec.~\ref{sec:main-analysis}.
Finally, we perform the ablation experiments in Sec.~\ref{sec:ablation-study}.

\subsection{Implementation}
\label{sec:setup}

\subsubsection{Datasets Preparation}
Our training dataset consists of synthetic sources and open sources.
The synthetic dataset is generated by Wan2.1-T2V-14B~\cite{wan2025wan}, where the captions are from OpenVID~\cite{nanopenvid}.
It is used for the distributional alignment during the distillation process.
The open datasets consist of the Koala-36M~\cite{wang2025koala} and the Intern4K~\cite{lin2025cascadev}. It is used to enhance the  upper bound of distillation performance.

\subsubsection{Training Configuration}
In the stability priming process, we train the model with a learning rate of $1\times 10^{-6}$ and a batch size of 32.
We use the Adam optimizer~\cite{kim2025adam} with $\beta=(0.9,0.999)$. This process is trained for only 500 steps.
In the adversarial process, we train the model with a larger learning rate of $2\times 10^{-6}$ and a batch size of 32.
We adopt the Adam optimizer with the smaller beta $\beta=(0.5,0.999)$ for lower memory consumption.
And we apply the EMA with a decay rate of 0.995.
For the generator, we set $\lambda=10$ to prevent the conditional frame collapse in I2V training.
For the discriminator heads, we set $\sigma_s=0.01$ and $\sigma_t=0.1$ to avoid the gradient explosion of the discriminator heads .
This process is trained for 1000 steps.

\subsubsection{Evaluation Metrics}
We primarily conduct the quantitative evaluation on the comprehensive benchmark VBench-I2V~\cite{huang2024vbench}.
However, we also provide the quantitative results for the traditional benchmark VFHQ~\cite{Xie_2022_VFHQ}.
The former includes 7 metrics and 3 dimensions: semantic alignment, temporal coherence, and frame quality.
The latter is used to measure the distributional distance between generated and real videos.
Additionally, we evaluate the diffusion latency of the inference process for all experiments on $8\times$H20 GPUs.

\subsubsection{Baseline Methods}
To validate the efficiency and quality of our approach, we compared it with other primary methods across different distillation paradigms.
It includes the Variational Score Distillation (VSD)~\cite{yin2024one,yin2024improved,shao2025magicdistillation}, the Consistency Distillation (CD)~\cite{song2023consistency,kimconsistency,lv2025dcm} and the Adversarial Distillation (AD)~\cite{lin2024sdxl,sauer2024fast,sauer2024adversarial,lin2025diffusion}.

\subsection{Main Results}
\label{sec:main-results}

\subsubsection{Quantitative Result}
We present the quantitative results on VBench-I2V in Tab.~\ref{tab:main_quantitative}.
The results demonstrate that V-PAE outperforms existing acceleration methods under an identical Number of Function Evaluations (NFE).
For 1-NFE, our approach outperforms the suboptimal method by an average of 5.8\% in the overall quality score, which includes semantic alignment, temporal coherence, and frame quality.
Compared to the 100-NFE baseline model, our approach achieves competitive performance in 1-NFE, which is only marginally behind by 1.5\% in the overall quality score. However, our approach in 4-NFE surpasses it by 3.3\%.
For the diffusion latency, our approach accelerates the diffusion process by $100\times$.
The more quantitative results on VFHQ can be found in the Appendix.


\subsubsection{Qualitative Result}
We present the qualitative comparison with existing acceleration methods in Fig.~\ref{fig:qualitative01}.
The results demonstrate that V-PAE produces clearer and more coherent videos in 1-NFE.
The comparison with the baseline model is shown in Fig.~\ref{fig:qualitative02}.
The baseline model produces videos with blurry frames, abrupt transitions between conditional and generated frames, and motion collapse due to unstable dynamic in 1-NFE.
In contrast, our approach generates consistently sharp and coherent frames.
The more qualitative results with other acceleration methods and the baseline model are also presented in the Appendix.


\subsection{Validating the Design}
\label{sec:main-analysis}
\begin{figure}[tb]
  \centering
  \includegraphics[width=0.85\columnwidth]{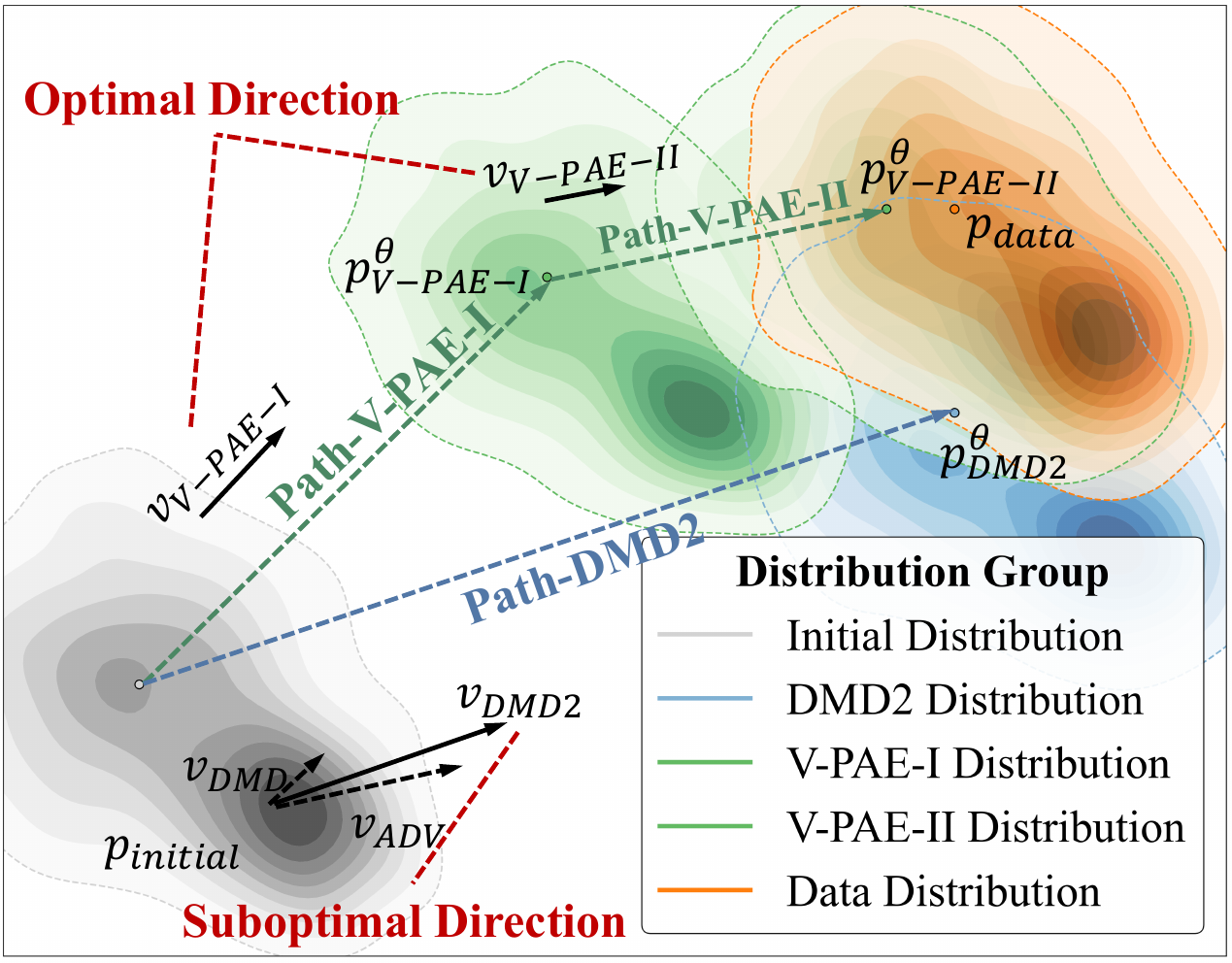}  
  \caption{
    \textbf{Fundamental analysis of V-PAE and DMD2}.
    V-PAE can optimizes the distribution along an optimal trajectory through phased optimization.
    DMD2 simultaneously moves along a suboptimal direction without error correction.
}
  \label{fig:analysis02}
\end{figure}

We primarily validate the necessity of our design based on the following analysis.
\textit{First, the role of  the stability priming process is confirmed through ablations:}
(i)Omitting this process entirely, (ii) implementing consistency distillation~\cite{song2023consistency} of APT~\cite{lin2025diffusion}, (iii) implementing our stability priming process.
The quantitative results in Tab.~\ref{tab:analysis} demonstrate that our design is well-motivated and necessary in the first process. And its positive advantages can be extended to the following process.
Notably, omitting the stability priming process leads to adversarial instability toward larger distribution biases.
\textit{Second, empirical observations reveal fundamental differences between V-PAE and DMD2.}
Our approach and DMD2 are both optimized through adversarial distillation and score sampling distillation. But there are fundamental differences between them.
The detailed analysis is illustrated in Fig.~\ref{fig:analysis02}.
DMD2 simultaneously optimizes the model using both the variational score gradient difference and the adversarial loss, which is a simple combination of different losses.
In fact, it can cause the model to move along a suboptimal direction in the distribution space, leading to a persistent distributional mismatch between real and generated distributions.
In contrast, our core motivation is to align the distributions of real and generated videos in the first process. It can improve the stability of single-step adversarial distillation in the following process.
Thus, we can achieve the high-quality adversarial equilibrium under the closer distributions.
\begin{figure}[tb]
  \centering
  \includegraphics[width=0.95\columnwidth]{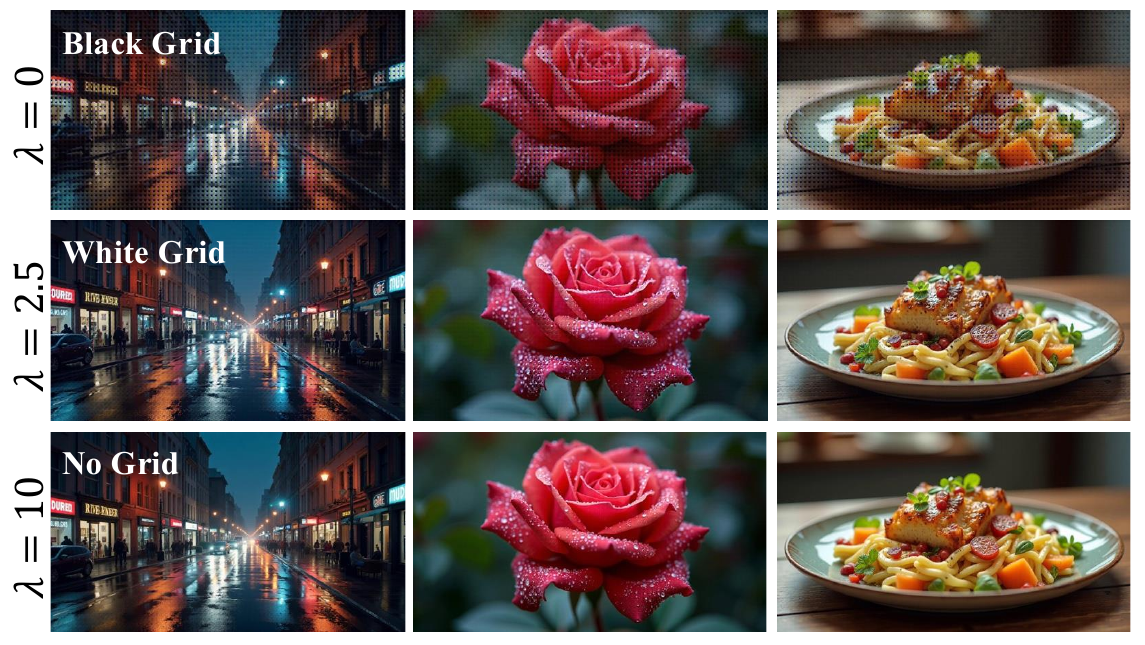}  
  \caption{
    Ablation study for the factor $\lambda$ of conditional Score Distillation Sampling loss. Zoom in for details.
  }
  \label{fig:ablation}
\end{figure}

\subsection{Ablation Study}
\label{sec:ablation-study}

\subsubsection{Discriminator Backbone Paradigm}
We compare the quantitative performance of the frozen, fully parameterized, and self discriminator in Tab.~\ref{tab:ablation}.
The self discriminator outperforms the other paradigms in the overall quality scores, demonstrating that the paradigm can identify more robust single-step video differences.
In addition, the fully parameterized training causes the Out-of-Memory (OOM) error, rendering it incompatible with the large-scale video model.

\begin{table}[tb]
\centering
\resizebox{0.95\columnwidth}{!}{%
\begin{tabular}{@{}l|c|c|c|c|c@{}}
\toprule
\textbf{Methods} &
\textbf{Type} &
\textbf{SA ($\uparrow$)} &
\textbf{TC ($\uparrow$)} &
\textbf{FQ ($\uparrow$)} &
\begin{tabular}[c]{@{}c@{}}\textbf{Quality} \\ \textbf{Score} ($\uparrow$)\end{tabular} \\
\midrule
\multicolumn{6}{c}{\textit{\textbf{First Process, 1-NFE}}} \\
\midrule
Baseline & Euler & 72.84 & 55.73 & 51.89 & 61.50 \\
LCM         & CD & 80.05 & 62.81 & 56.55 & 67.96 \\
V-PAE-I    & VSD & \textbf{84.92} & \textbf{65.83} & \textbf{59.75} & \textbf{72.34} \\
\midrule
\multicolumn{6}{c}{\textit{\textbf{Second Process, 1-NFE}}} \\
\midrule
UAE + Baseline   & AD & 83.12 & 63.58 & 57.37 & 69.65 \\
UAE + LCM        & AD & 87.94 & 69.62 & 62.51 & 75.21 \\
UAE + V-PAE-I    & AD & \textbf{94.97} & \textbf{72.26} & \textbf{65.47} & \textbf{79.56} \\
\bottomrule
\end{tabular}
}
\caption{
\textbf{Necessity of stability priming.} 
The comparison of different first-process initialization methods and their effects on subsequent adversarial training.
}
\label{tab:analysis}
\end{table}

\begin{table}[htb]
\centering
\resizebox{0.95\columnwidth}{!}{%
\begin{tabular}{@{}l|c|c|c|c|c@{}}
\toprule
\multirow{2}{*}{\textbf{Setting}} &
\multirow{2}{*}{\textbf{SA} ($\uparrow$)} &
\multirow{2}{*}{\textbf{TC} ($\uparrow$)} &
\multirow{2}{*}{\textbf{FQ} ($\uparrow$)} &
\textbf{Quality} &
\multirow{2}{*}{\textbf{Memory}} \\
& & & &
\textbf{Score} ($\uparrow$) &
\\
\midrule
\multicolumn{6}{c}{\textit{\textbf{Discriminator Head Architecture}}}  \\
\midrule
Conv     & 75.55 & 57.52 & 52.11 & 63.29 & IM \\
CA       & 91.36 & 72.23 & 65.29 & 78.31 & IM \\
CA$^\dagger$        & \textbf{94.97} & \textbf{72.26} & \textbf{65.47} & \textbf{79.56} & IM \\
\midrule
\multicolumn{6}{c}{\textit{\textbf{Discriminator Backbone Paradigm}}} \\
\midrule
Frozen  & 92.21 & 70.12 & 63.92 & 77.36 & IM  \\
Trained & -     & -     & -     & -     & OOM \\
Unified  & \textbf{94.97} & \textbf{72.26} & \textbf{65.47} & \textbf{79.56} & IM  \\ 
\midrule
\multicolumn{6}{c}{\textit{\textbf{Factor of Conditional SDS $\lambda$}}}  \\
\midrule
$\lambda=0$     & 80.36 & 61.15 & 55.41 & 67.33 & IM \\
$\lambda=2.5$   & 86.83 & 66.18 & 59.96 & 72.87 & IM \\
$\lambda=10$    & \textbf{94.97} & \textbf{72.26} & \textbf{65.47} & \textbf{79.56} & IM \\
$\lambda=100$   & 92.28 & 70.24 & 63.62 & 77.31 & IM \\
\bottomrule
\end{tabular}
}
\caption{
\textbf{Component-wise ablation study.} 
Videos are generated in 1-NFE. Memory is measured on $32\times$H20 GPUs. 
Abbreviations -- SA: Semantic Alignment; TC: Temporal Coherence; FQ: Frame Quality; CA$^{\dagger}$: Multi-modal Cross-Attention; OOM: Out of Memory; IM: In Memory.
}
\label{tab:ablation}
\end{table}

\subsubsection{Discriminator Head Architecture}
We compare the quantitative results for the different discriminator head architectures in Tab.~\ref{tab:ablation}.
The results show that the design of the head architecture has a significant impact on semantic alignment.
For the large-scale video model, the convolution head is unable to stabilize adversarial training,which leads to a degradation in the overall score quality.

\subsubsection{Factor of Conditional SDS}
We ablate the factor of the conditional SDS loss in Tab.~\ref{tab:ablation}.
When $\lambda=0$, the conditional and generated frames suffer from severe collapse and discontinuities, as shown in Fig.~\ref{fig:ablation}.
As the $\lambda \rightarrow 10$ , the overall quality score has significantly improved. And the conditional frame collapse is gradually restored.
As the $\lambda \rightarrow 100$, the video sharpness is slightly impacted, and the frame quality score is reduced.

\section{Conclusion}
\label{sec:conclusion}
In this paper, we propose the Video Phased Adversarial Equilibrium (V-PAE), a distillation framework that enables high-quality, single-step video generation from large-scale video models. Our approach employs a two-phase process.
(i) Stability priming is a warm-up process designed to improve the stability of single-step adversarial distillation in the following process.
(ii) Unified adversarial equilibrium is a flexible self-adversarial process that achieves a co-evolutionary adversarial equilibrium in the Gaussian noise space.
We also primarily preserve video-image subject consistency in image-to-video (I2V) generation.
Comprehensive experiments on VBench-I2V demonstrate that V-PAE outperforms existing acceleration methods by an average of 5.8\% in the overall quality score.
And our approach reduces the diffusion latency of the large-scale video model by $100\times$, while preserving competitive performance.

\bibliography{aaai2026}

\begin{appendices}
\setlength{\textfloatsep}{16pt plus 2pt minus 4pt}
\setlength{\floatsep}{16pt plus 2pt minus 2pt}
\setlength{\intextsep}{4pt plus 2pt minus 2pt}
\setlength{\abovecaptionskip}{5pt}
\setlength{\belowcaptionskip}{0pt}
\setlength{\dbltextfloatsep}{4pt plus 2pt minus 4pt}
\setlength{\dblfloatsep}{4pt plus 2pt minus 2pt}

\setlength{\abovedisplayskip}{10pt plus 2pt minus 5pt}
\setlength{\belowdisplayskip}{10pt plus 2pt minus 5pt}
\setlength{\abovedisplayshortskip}{0pt plus 3pt}
\setlength{\belowdisplayshortskip}{6pt plus 3pt minus 3pt}

\clearpage

\twocolumn[
\begin{center}
    \LARGE
    \textbf{Phased One-Step Adversarial Equilibrium for Video Diffusion Models}\\
    \vspace{0.5em}
    \Large
    \textbf{Appendix}\\
    \vspace{1.0em}
\end{center}
]

We organize the appendix as follows:
\begin{itemize}
\setlength{\itemsep}{0pt}
\setlength{\parsep}{0pt}
    \item Appendix A\quad Mathematical Preliminaries
    \item Appendix B\quad Experiment Details
    \item Appendix C\quad Additional Results

\end{itemize}

\section{Mathematical Preliminaries}
\subsection{Diffusion Models}
\label{sec:diffusion_model}

Diffusion models~\cite{ho2020denoising,songscore,songdenoising} constitute a class of generative models that learn to synthesize data by approximating the reverse process of a stochastic forward transformation that progressively corrupts data into noise. 
The mathematical foundation relies on defining a forward process that transforms the data distribution $x_0 \sim p_\text{data}$ into a noise distribution $x_T \sim \mathcal{N}(0,\textbf{I})$ and subsequently learns a model that approximates the reverse process $p(x_{t-1}|x_t)$ to progressively denoise samples from pure noise to data.
From an Ordinary Differential Equation (ODE) modeling perspective, diffusion processes can be understood through continuous-time Probability Flow (PF). 
Researchers formulate the denoising process as learning a deterministic PF ODE $dx_t = v(x_t,t) dt$ that transforms the noise distribution $p(x_T)$ back to $p(x_0)$ which ideally approximates $p_{\text{data}}(x_0)$.
By learning $v(x_t,t)$, this enables sampling according to $p_\text{data}(x_0)$ by first sampling $\epsilon \sim \mathcal{N}(0,\textbf{I})$ and then using ODE solvers $\mu^\theta$ to numerically integrate the trajectory of the PF ODE to arrive at $p(\hat{x}^\theta_0)$.
Where the forward process follows a Gaussian noise injection paradigm, the velocity field $v(x_t, t)$ is parameterized as:
\begin{equation}
v(x_t, t) = f(t) x_t  - \frac{1}{2} g^2(t) s(x_t, t),
\end{equation}
where $f(t), g(t)$ are the predetermined drift and diffusion coefficients determined by the noise schedule, and $s(x_t, t) = \nabla_{x_t} \log p(x_t)$. 
Alternatively, if we consider the linear interpolation~\cite{lipman2022flow}, whose forward process is defined as:
\begin{equation}
x_t = (1-t)x_0 + t\epsilon,
\end{equation}
between the data sample $x_0$ and Gaussian noise $\epsilon$. 
This formulation intuitively enables learning straight ODE trajectories that can be simulated with fewer integration steps compared to curved paths. 
If we parameterize the velocity by $v(x_t, t)=\mu^\theta(x_t, t)$, we can approximate $\frac{dx_t}{dt} = \epsilon-x_0$ by optimizing the following objective:
\begin{equation}
\mathcal{L} = \mathbb{E}_{\substack{x_0 \sim p_{\text{data}},\\ \epsilon \sim \mathcal{N}(0,\textbf{I})}} \left[\|(\epsilon-x_0)-\mu_\theta(x_t,t) \|^2\right].
\end{equation}

\subsection{Variational Score Distillation}
Distribution Matching Distillation (DMD)~\cite{yin2024one} is an instance of Variational Score Distillation (VSD)~\cite{pooledreamfusion}.
Its objective is to distill a single-step generator $\mu^{\theta}$ that produces $\hat{x}^\theta_0$ from Gaussian noise $\epsilon$. 
This is achieved by minimizing the KL divergence between the distributions $p_{\text{real}}(\hat{x}^\theta_0)$ and $p^\phi_\text{fake}(\hat{x}^\theta_0)$, which are governed by the diffusion models  $\mu_{real}$ and $\mu^\phi_\text{fake}$.
The gradient update of this KL divergence is given by
\begin{align*}
&\frac{\partial}{\partial \theta}\mathcal{D}_\text{KL}(p_{\mu_\text{real}}(\hat{x}^\theta_0) \| p^\phi_\text{fake}(\hat{x}^\theta_0)) \\
=& \frac{\partial}{\partial \theta}\mathbb{E}_{\epsilon \sim \mathcal{N}(0,\bf{I}), \hat{x}_0^\theta=\epsilon-T\mu^\theta(\epsilon)}\log\left(  \frac{p_{\text{real}}(\hat{x}^\theta_0)}{p^\phi_\text{fake}(\hat{x}^\theta_0)} \right)\\
=& \mathbb{E}_{\epsilon \sim \mathcal{N}(0,\bf{I}), \hat{x}_0^\theta=\epsilon-T\mu^\theta(\epsilon)}\left( s_{\text{real}}(\hat{x}^\theta_0) -s^\phi_\text{fake}(\hat{x}^\theta_0) \right)\frac{\partial \hat{x}^\theta_0}{\partial \theta},
\end{align*}
where $s_\text{real}(\hat{x}^\theta_0)=\nabla_{\hat{x}^\theta_0}\log p_{\text{real}}(\hat{x}^\theta_0)$ and $s^\phi_\text{fake}(\hat{x}^\theta_0)=\nabla_{\hat{x}^\theta_0}\log p^\phi_\text{fake}(\hat{x}^\theta_0)$ are the score functions.
In practice, the single-step generator $\mu^\theta$ is updated using this distribution matching loss, while the fake model $\mu^\phi_\text{fake}$ is updated alternately.
This alternating schedule ensures that $\mu^\phi_\text{fake}$ continuously adapts to the distribution of $\mu^\theta$ throughout training.

\section{Experiment Details}
\subsection{Training Configuration}
Our comparable acceleration methods include the Latent Consistency Model (LCM)~\cite{luo2023latent} and Phased Consistency Model (PCM)~\cite{wang2024phased} from Consistency Distillation (CD), DMD2~\cite{yin2024improved}  and MagicDistillation~\cite{shao2025magicdistillation} from Variational Score Distillation (VSD), Adversarial Diffusion Distillation (ADD)~\cite{sauer2024fast}  and Adversarial Post-Training (APT) from Adversarial Distillation (AD).
To preserve fairness among comparable acceleration methods, we conduct experiments using the same model~\cite{wan2025wan}, dataset, and training cost. 
To facilitate reproducibility, we provide detailed hyperparameter settings for comparable acceleration methods during distillation training.

\subsubsection{LCM and PCM}
For LCM, We employ the student model $\mu^\theta$ and the teacher model $\mu_\text{tea}$ from the same initialized parameters.
In the trajectory of the PF ODE, we sample 64 points for $\mu_\text{tea}$, guiding the $\mu^\theta$ to map towards the boundary points (\eg, $t=0$). 
We set the batch size to 32, and the learning rate is fixed at $1\times10^{-6}$.
We enable the Exponential Moving Average (EMA) with a decay rate of 0.95 to improve training stability.
We set the factor of the Huber loss to 0.001.
For PCM, we additionally incorporate a simple discriminator like ~\cite{wang2024phased} to improve the frame quality of videos.

\subsubsection{DMD2 and MagicDistillation}
For DMD2, we employ the student model $\mu^\theta$, the real model $\mu_\text{real}$, and the fake model $\mu^\phi_\text{fake}$ with the same initialized parameters.
The $\mu^\theta$ and $\mu^\phi_\text{fake}$ are updated at a ratio of $1:5$.
We sample the Gaussian noise $\epsilon$ into $\mu^\theta$ to obtain $\hat{x}^\theta_0$, which is perturbed by varying noise levels from the interval $(0,T]$ to $\mu_\text{real}$ and $\mu^\phi_\text{fake}$.
We set the batch size to 32, and the learning rate is fixed at $1\times10^{-6}$.
For the distillation loss, we use the $\mu^\phi_\text{fake}$ as the discriminator backbone for feature extraction.
For MagicDistillation, we additionally introduce the Low-Rank Adaptation (LoRA)~\cite{hulora} to the $\mu_\text{real}$ and $\mu^\phi_\text{fake}$, and further incorporate a supervised loss.

\subsubsection{ADD and APT}
For ADD, we employ the student model $\mu^\theta$, the teacher model $\mu_\text{tea}$ and a discriminator head $d^\psi$.
We set the frozen $\mu_\text{tea}$ as the discriminator backbone, with a convolutional head.
The $\mu^\theta$ performs adversarial training at four uniform points of the PF ODE trajectory $[T/4,T/2,3T/4,T]$.
We set the batch size to 32, and the learning rate is fixed at $1\times10^{-6}$.
We enable the EMA with a larger decay rate of 0.995 to avoid instability in adversarial performance.
For APT, we first train $\mu^\theta$ with LCM, and perform adversarial training at $T=1$.
For the large-scale video model (Wan2.1-I2V-14B)~\cite{wan2025wan}, fully-parameterized training for the discriminator is infeasible due to resource constraints.
Therefore, we use $\mu_\text{tea}$ as the discriminator backbone and incorporate the cross-attention discriminator head, like~\cite{lin2025diffusion}.
The above experiments are trained for 1,000 iterations.

\subsection{Inference Setup}

We generate the videos without any guidance scale for distilled models.
Its sampling step is equivalent to the Number of Function Evaluation (NFE).
For the baseline model, we generate videos with a guidance scale of 5, and its sampling step is twice as long as NFE.
We set the flow shift to 5.0 for every distilled model and the baseline model, and then employ the Euler solver to sample.

\subsection{Evaluation Metrics}
We use the two primary evaluation metrics for the video model, which are the paradigms from the traditional distribution distance and comprehensive multi-modal metrics.

\subsubsection{Distribution Distance}
The traditional metrics evaluate video quality by distribution distance,  referred to as Fréchet Inception Distance (FID)~\cite{FID} and Fréchet Video Distance (FVD)~\cite{unterthiner2019fvd}.
Following the protocol of MagicDistillation~\cite{shao2025magicdistillation}, we randomly select 100 video clips from the VFHQ~\cite{Xie_2022_VFHQ}.
Each clip consists of a 5-second video at a resolution of $540\times960$.
We compute the FID and FVD by comparing the generated videos with the corresponding real videos, thereby quantitatively evaluating video quality.

\subsubsection{Multi-modal Metrics}
The traditional metrics have clear limitations for high-quality video generation.
With recent progress, video models can surpass real videos in terms of aesthetics and visual quality.
It renders the distribution distance inadequate for a holistic evaluation.
Comprehensive VBench-I2V~\cite{huang2024vbench} addresses this gap.
It is a mainstream benchmark for large-scale video diffusion models that combines automated evaluation with human annotations instead of relying solely on distributional distances.
This assesses video models in the I2V task along multiple axes, including semantic alignment, temporal coherence, and frame quality.
Specifically, we follow the VBench-I2V protocol. 
We generate videos using the official prompts and images, producing five videos per prompt–image pair with different random seeds. 
We then apply a multimodal understanding model augmented with human feedback to compute fine-grained scores: overall I2V score; semantic alignment (subject and background consistency); temporal coherence (motion smoothness and dynamics); and frame quality (aesthetics and image fidelity). 
This setup provides a more comprehensive and nuanced assessment of generated videos.


\section{More Results}
\subsection{Quantitative Results}

\begin{table}[tb]
\centering
\resizebox{0.8\columnwidth}{!}{
\begin{tabular}{@{}l|c|c|c@{}}
\toprule
\textbf{Methods}           & \textbf{Type} & \textbf{FID} ($\downarrow)$ & \textbf{FVD} ($\downarrow$) \\ \midrule
\multicolumn{4}{@{}c}{\textbf{\textit{100-NFE}}} \\
\midrule
Baseline               & Euler   &  31.89   &  172.13   \\
\midrule
\multicolumn{4}{@{}c}{\textbf{\textit{4-NFE}}} \\
\midrule
Baseline               & Euler   &   40.23  &  328.47   \\
PCM               & CD   &  38.76   &  287.62   \\
DCM               & CD   &  37.83   &  276.90   \\
DMD2              & VSD   &  33.87   &  218.78   \\
MagicDistillation & VSD   &   31.79  &  167.34   \\
ADD               & AD   &  35.51   &  269.67   \\
APT               & AD   &  32.76   &  184.87   \\
\textbf{V-PAE}       & AD   &   \textbf{31.13}  &  \textbf{162.80}   \\
\midrule
\multicolumn{4}{@{}c}{\textbf{\textit{1-NFE}}} \\
\midrule
APT               & AD   &  34.21   &  212.59   \\
\textbf{V-PAE}       & AD   &  \textbf{33.76}   &  \textbf{194.45}   \\ \bottomrule
\end{tabular}
}
\caption{
\textbf{Quantitative results} for I2V comparison on VFHQ. Baseline represent Wan2.1-I2V-14B~\cite{wan2025wan}. We sample videos of $720\times1280$ and reshape to $540\times960$ for evaluation.
}
\label{tab:vfhq_fvd}
\end{table}

Beyond VBench-I2V~\cite{huang2024vbench}, we also evaluate the distilled models on VFHQ~\cite{Xie_2022_VFHQ}.
As shown in Tab.~\ref{tab:vfhq_fvd}, V-PAE achieves state-of-the-art single-step performance, even surpassing the multi-step sampling of comparable acceleration distillation methods.
In the zero-shot manner, our approach of multi-step sampling exceeds the baseline model using 100-NFE sampling.
It further confirms the effectiveness of our approach.

\subsection{Qualitative Results}
\begin{figure*}[htb]
  \centering
  \includegraphics[width=0.95\linewidth]{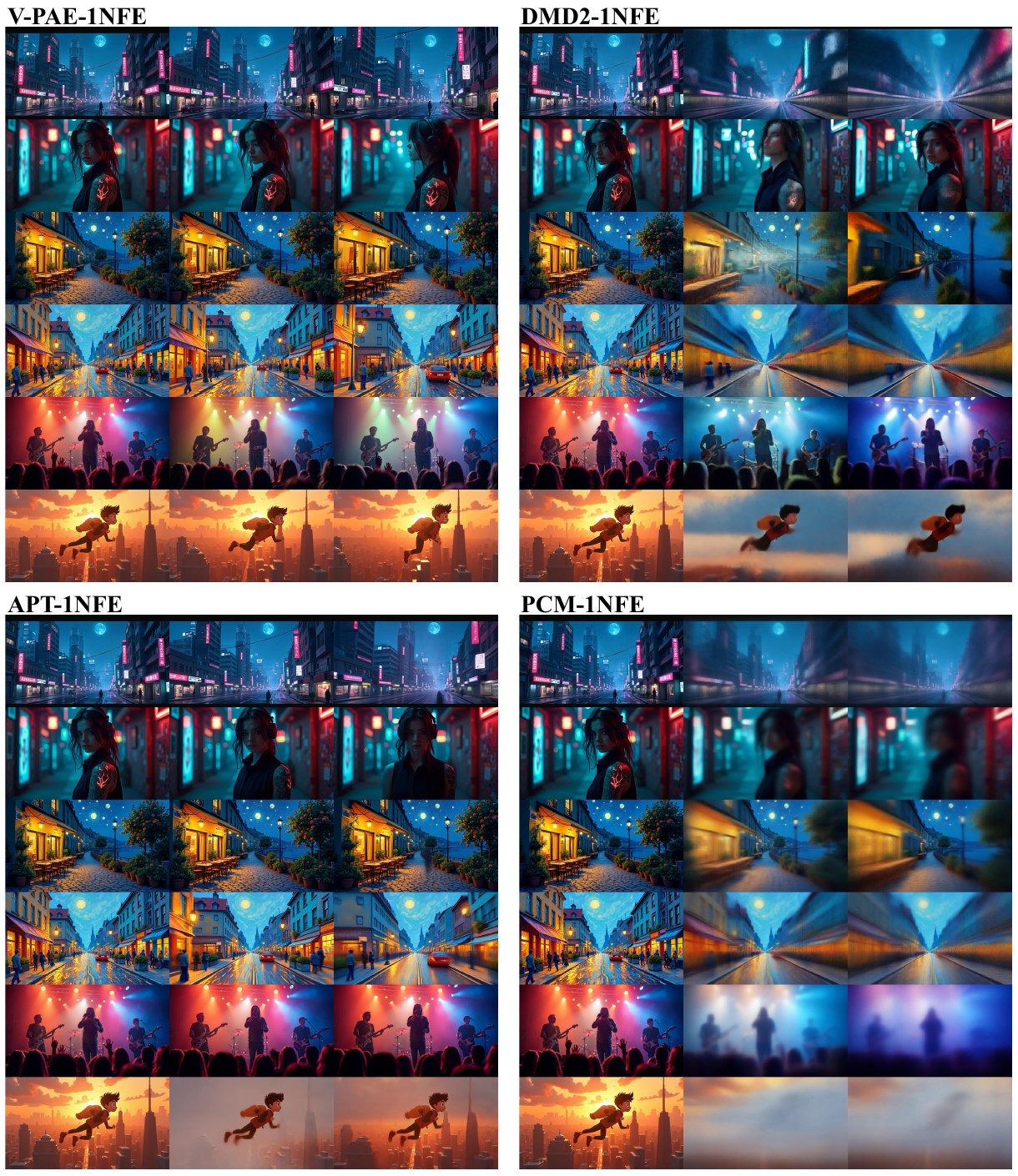}  
  \caption{
\textbf{Qualitative comparison of V-PAE with acceleration distillation methods. }
We evaluate V-PAE against three representative paradigms: DMD2 (VSD), PCM (CD), and APT (AD).
}
  \label{fig:app02}
\end{figure*}

\begin{figure*}[htb]
  \centering
  \includegraphics[width=0.95\linewidth]{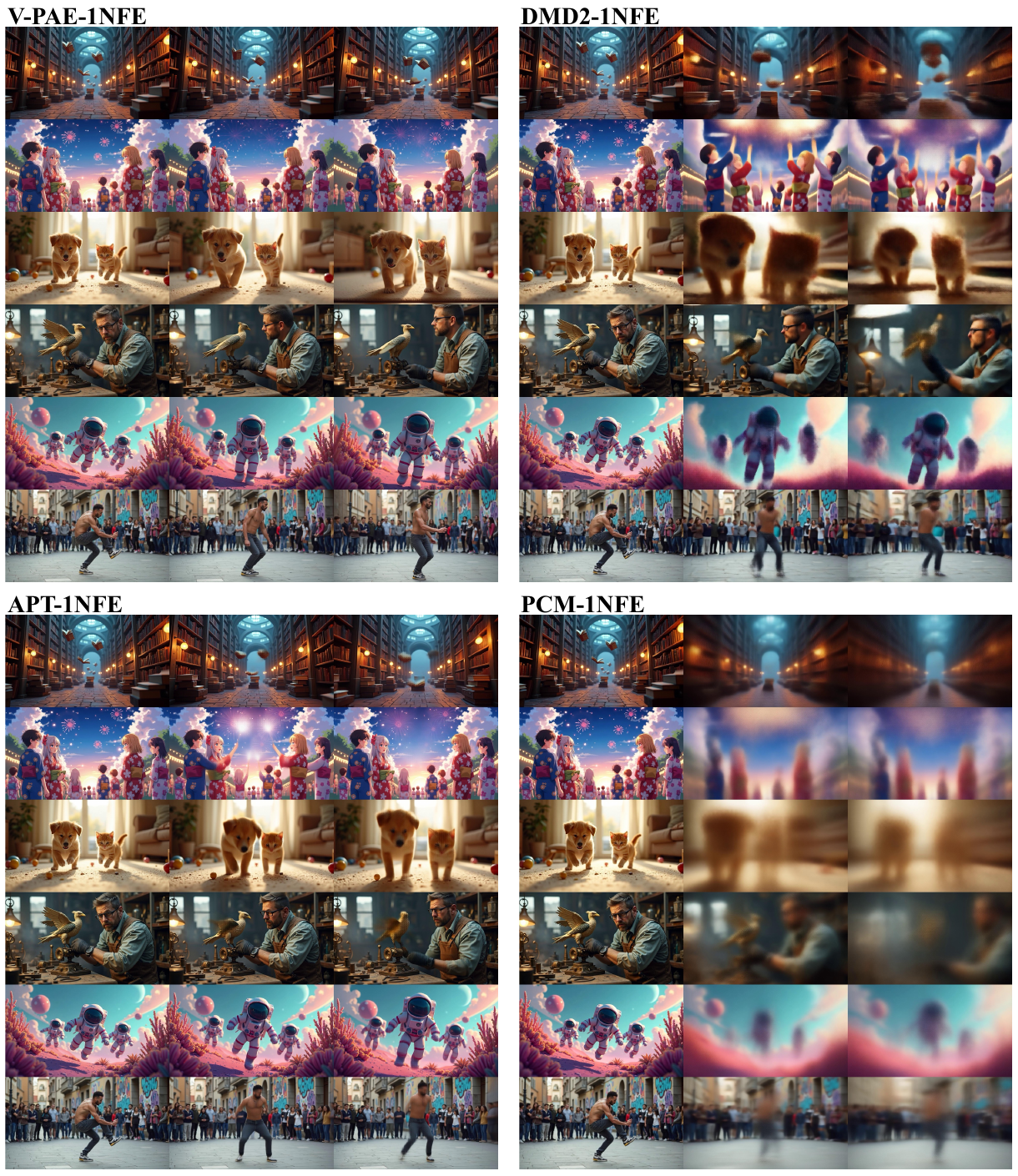}  
  \caption{
\textbf{Qualitative comparison of V-PAE with acceleration distillation methods. }
We evaluate V-PAE against three representative paradigms: DMD2 (VSD), PCM (CD), and APT (AD).
  }
  \label{fig:app03}
\end{figure*}

\begin{figure*}[htb]
  \centering
  \includegraphics[width=0.95\linewidth]{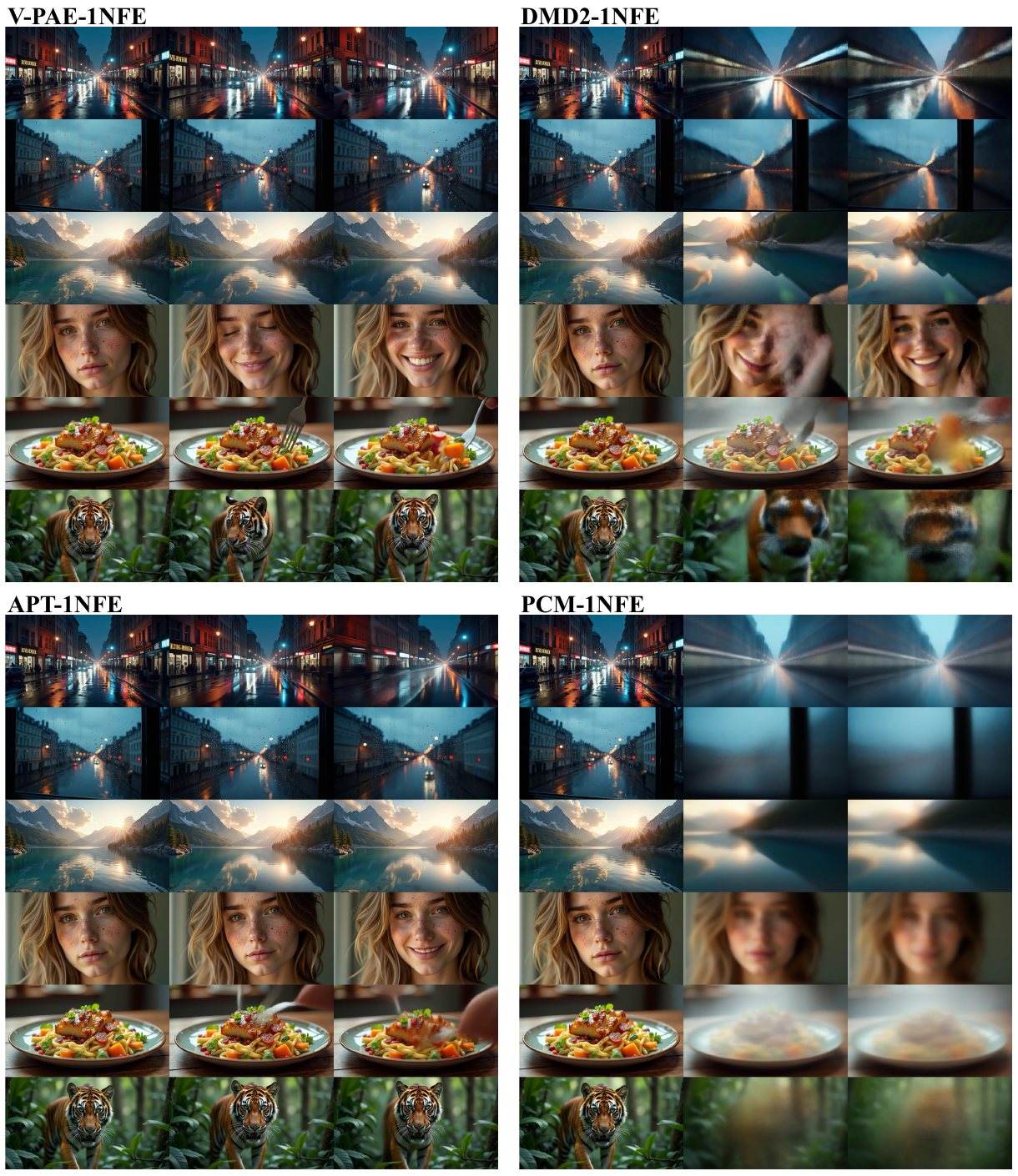}  
  \caption{
\textbf{Qualitative comparison of V-PAE with acceleration distillation methods. }
We evaluate V-PAE against three representative paradigms: DMD2 (VSD), PCM (CD), and APT (AD).
  }
  \label{fig:app04}
\end{figure*}

\begin{figure*}[htb]
  \centering
  \includegraphics[width=0.95\linewidth]{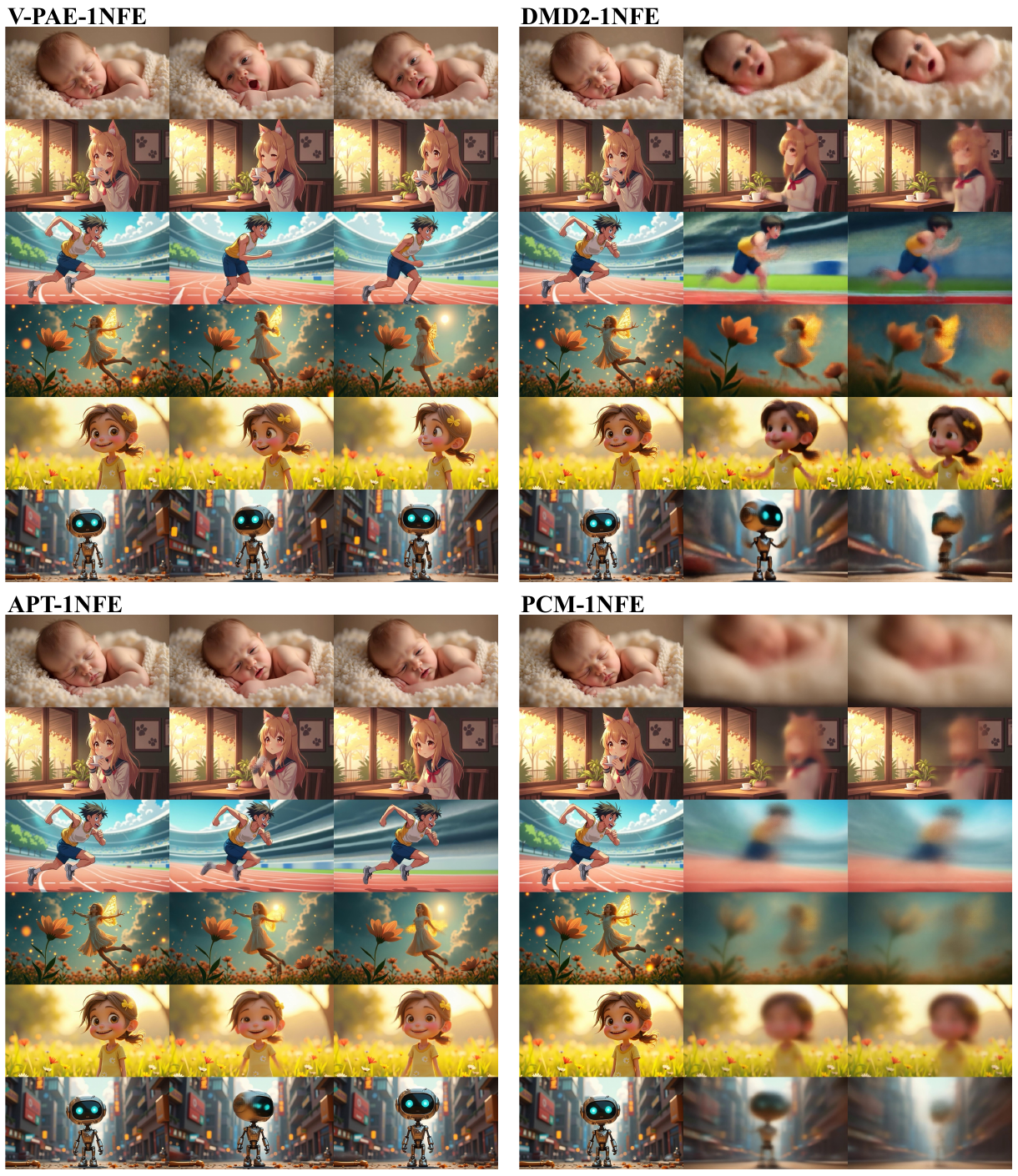}  
  \caption{
\textbf{Qualitative comparison of V-PAE with acceleration distillation methods. }
We evaluate V-PAE against three representative paradigms: DMD2 (VSD), PCM (CD), and APT (AD).
  }
  \label{fig:app05}
\end{figure*}

\begin{figure*}[htb]
  \centering
  \includegraphics[width=0.95\linewidth]{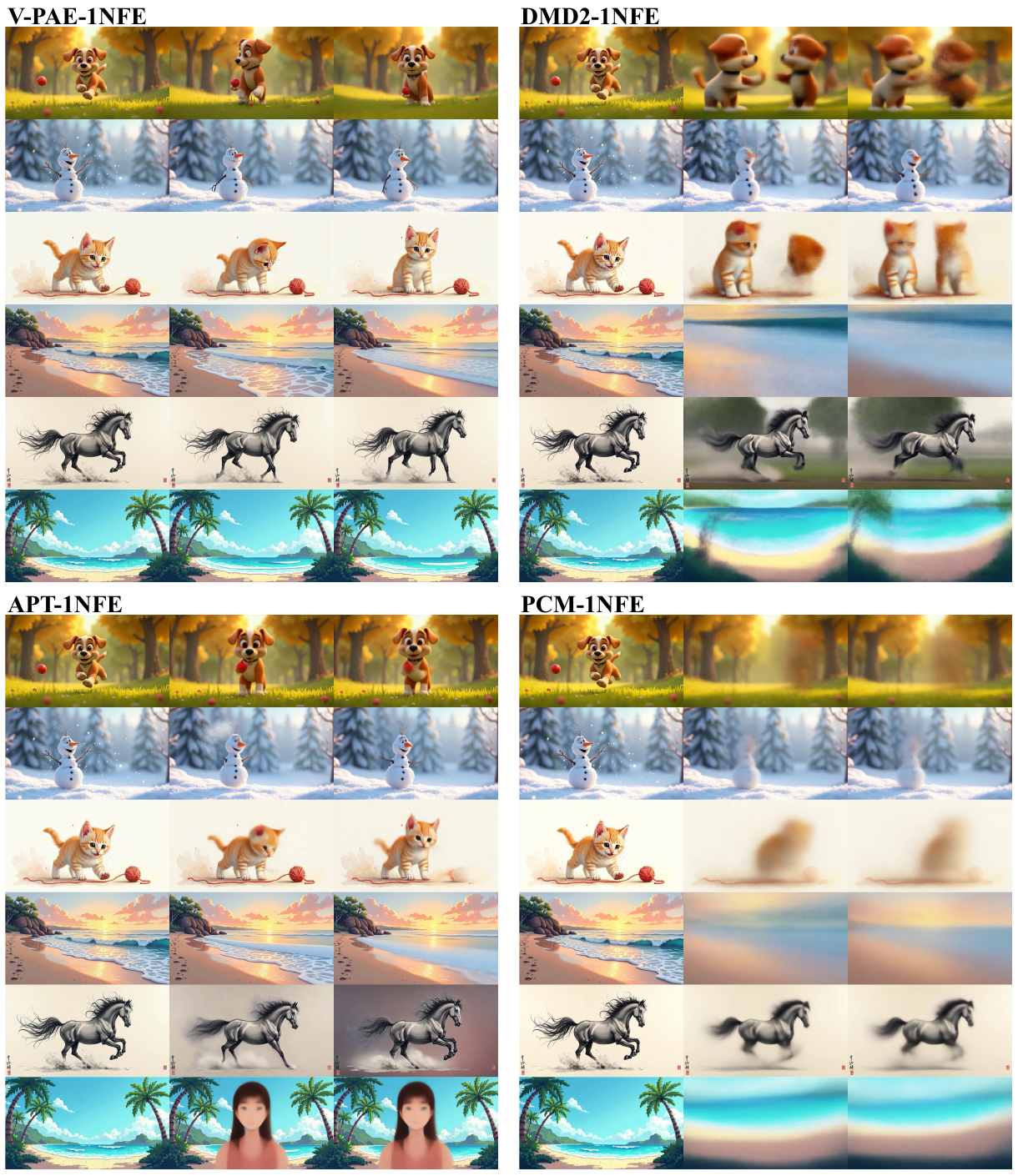}  
  \caption{
\textbf{Qualitative comparison of V-PAE with acceleration distillation methods. }
We evaluate V-PAE against three representative paradigms: DMD2 (VSD), PCM (CD), and APT (AD).
  }
  \label{fig:app06}
\end{figure*}

\begin{figure*}[htb]
  \centering
  \includegraphics[width=0.9\linewidth]{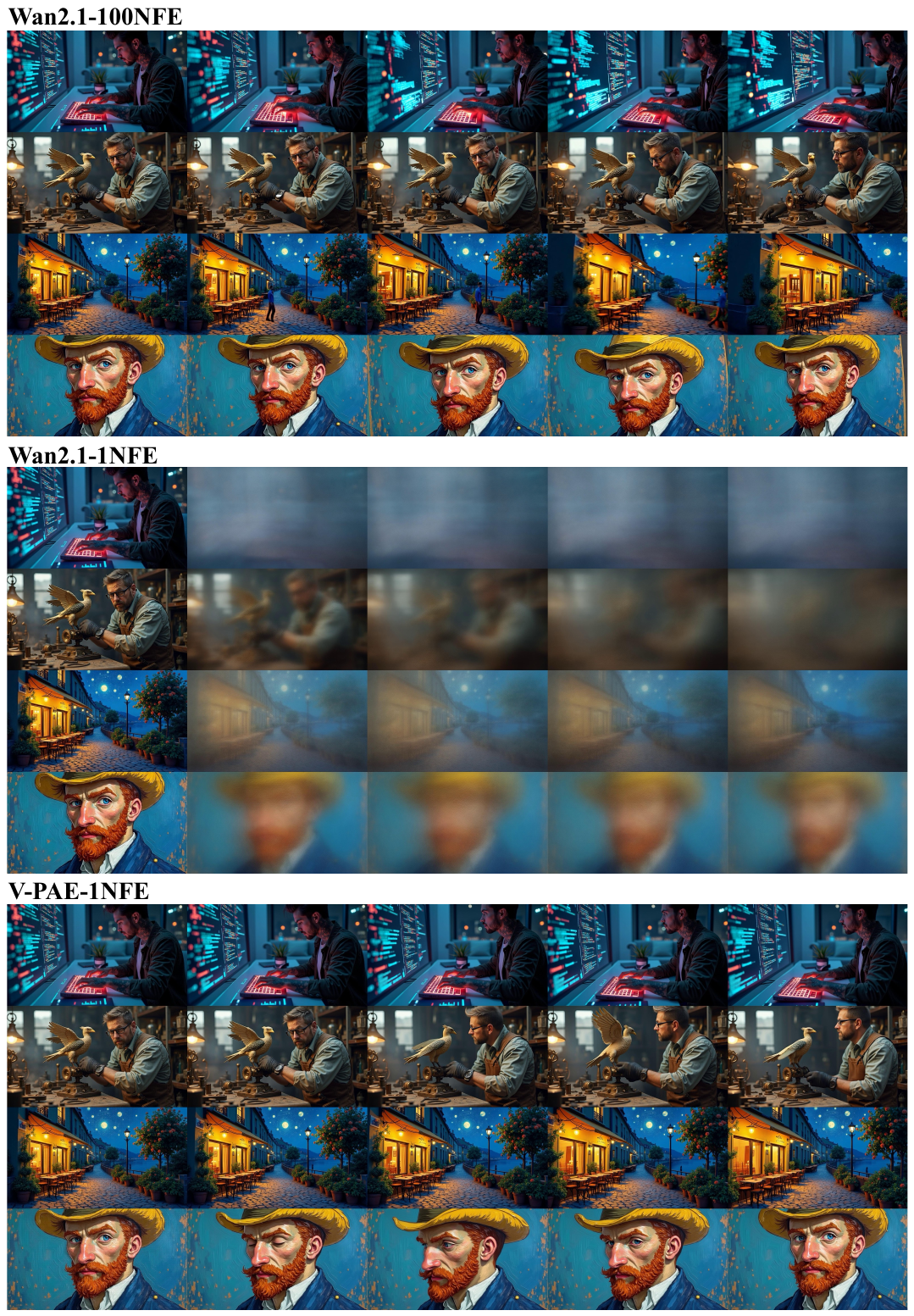}
  \caption{
\textbf{Qualitative comparison of V-PAE with Wan2.1-I2V-14B. }
We generate videos from the baseline under two sampling regimes: 100 NFE and 1 NFE. 
For the 100-NFE setting, we use 50 sampling steps with an guidance scale.
  }
  \label{fig:app07}
\end{figure*}

\begin{figure*}[htb]
  \centering
  \includegraphics[width=0.9\linewidth]{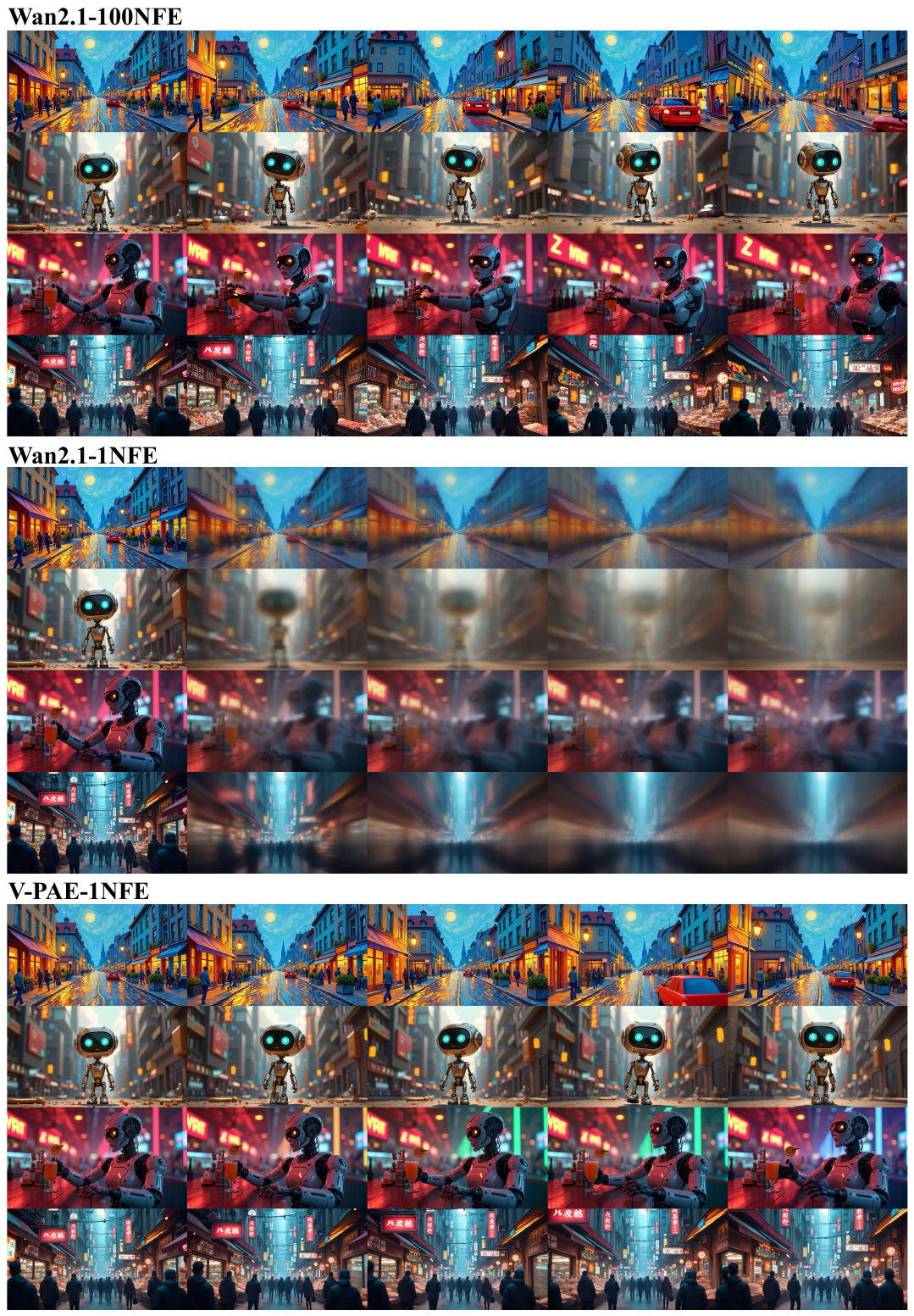}
  \caption{
\textbf{Qualitative comparison of V-PAE with Wan2.1-I2V-14B. }
We generate videos from the baseline under two sampling regimes: 100 NFE and 1 NFE. 
For the 100-NFE setting, we use 50 sampling steps with an guidance scale.
  }
  \label{fig:app08}
\end{figure*}

\begin{figure*}[htb]
  \centering
  \includegraphics[width=0.9\linewidth]{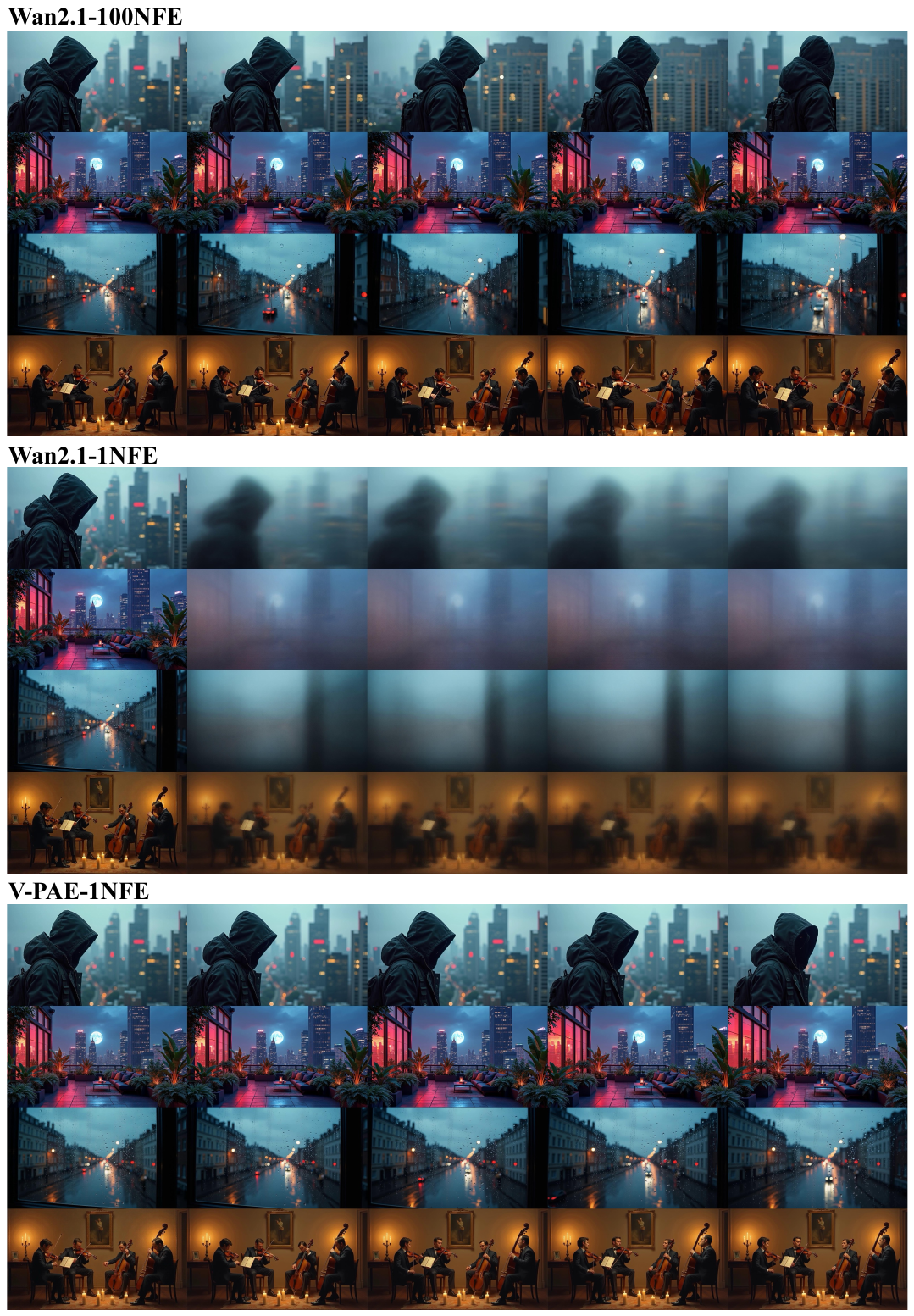}
  \caption{
\textbf{Qualitative comparison of V-PAE with Wan2.1-I2V-14B. }
We generate videos from the baseline under two sampling regimes: 100 NFE and 1 NFE. 
For the 100-NFE setting, we use 50 sampling steps with an guidance scale.
  }
  \label{fig:app09}
\end{figure*}

\begin{figure*}[htb]
  \centering
  \includegraphics[width=0.9\linewidth]{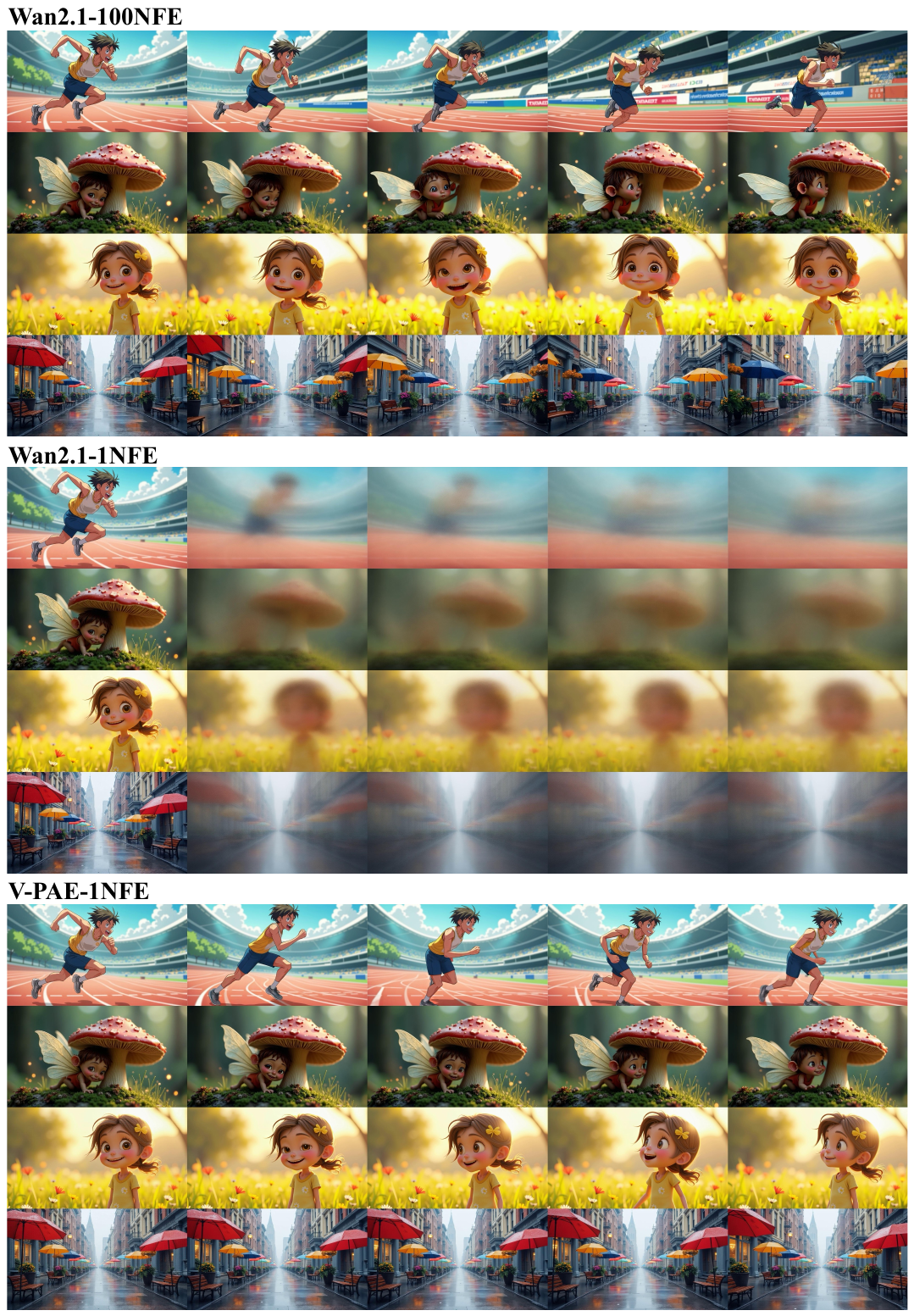}
  \caption{
\textbf{Qualitative comparison of V-PAE with Wan2.1-I2V-14B. }
We generate videos from the baseline under two sampling regimes: 100 NFE and 1 NFE. 
For the 100-NFE setting, we use 50 sampling steps with an guidance scale.
  }
  \label{fig:app10}
\end{figure*}

\begin{figure*}[htb]
  \centering
  \includegraphics[width=0.9\linewidth]{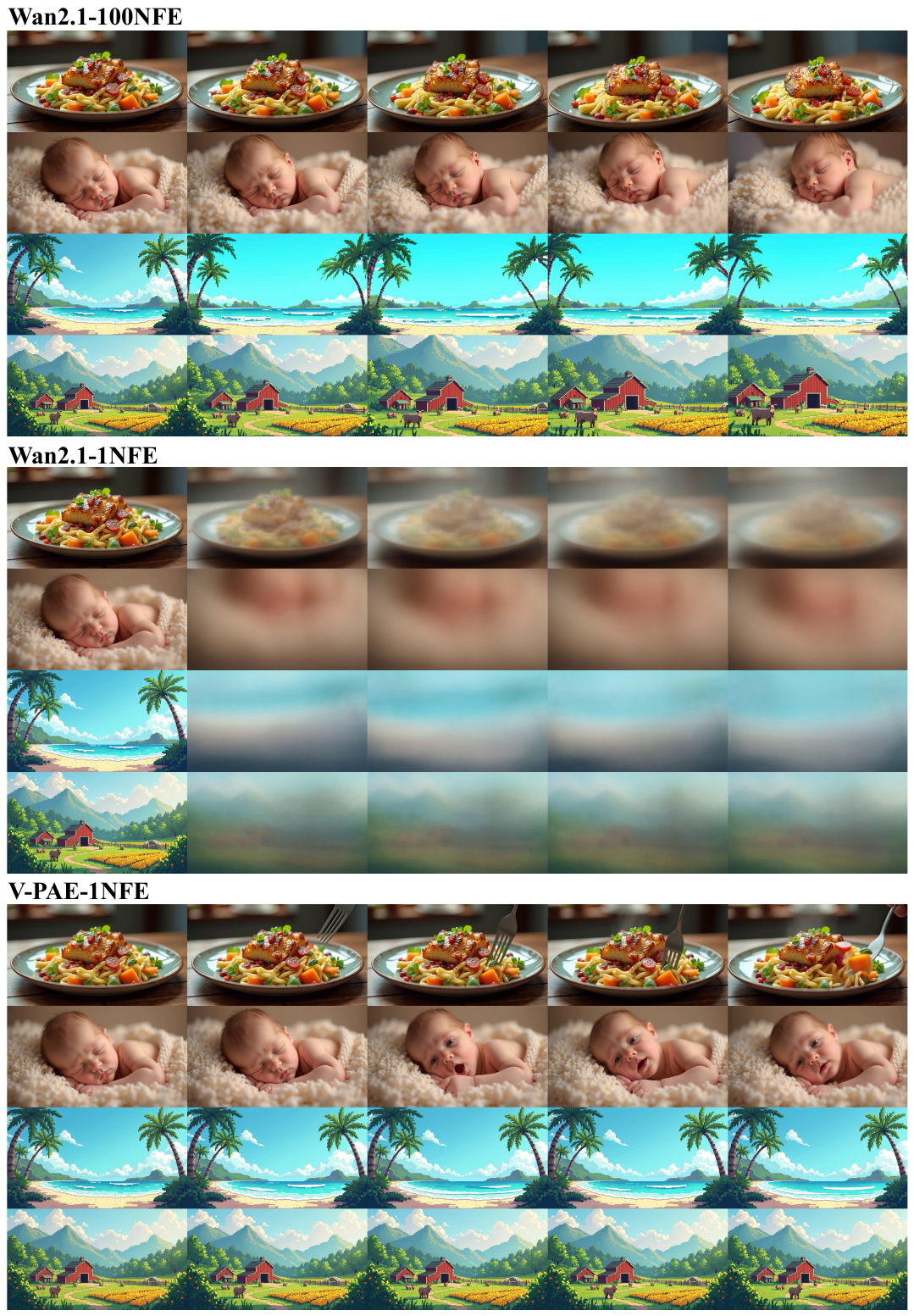}
  \caption{
\textbf{Qualitative comparison of V-PAE with Wan2.1-I2V-14B. }
We generate videos from the baseline under two sampling regimes: 100 NFE and 1 NFE. 
For the 100-NFE setting, we use 50 sampling steps with an guidance scale.
  }
  \label{fig:app11}
\end{figure*}

We provide more qualitative results for the visual comparison with acceleration distillation methods, which are shown in Figs.~\ref{fig:app02}-\ref{fig:app06}.
We also provide more qualitative results for the visual comparison with the baseline model, which are shown in Figs.~\ref{fig:app07}-\ref{fig:app11}.
The images and prompts used for I2V generation are produced by FLUX.1-dev~\cite{batifol2025flux} and GPT-4~\cite{achiam2023gpt}.


\end{appendices}

\end{document}